\documentclass{article}

\usepackage[margin=1in]{geometry}
\usepackage{wrapfig}
\usepackage{multirow}
\usepackage{adjustbox}
\usepackage{graphicx}
\usepackage{booktabs}
\usepackage{hyperref}
\usepackage{natbib}
\usepackage{xcolor}
\usepackage{amsfonts}       
\usepackage{microtype}
\usepackage{graphicx}
\usepackage{subcaption}

\newcommand{\ictc}{\textsc{IC\textbar TC}}
\usepackage[table, svgnames, dvipsnames]{xcolor}
\usepackage{colortbl}
\definecolor{lightblue}{RGB}{198, 211, 231}
\usepackage{makecell, cellspace, caption}
\usepackage{amsmath}
\usepackage{enumitem}
\usepackage{wrapfig}
\newif\ifarxiv
\arxivtrue
\hypersetup{
    colorlinks=true,
    linkcolor=blue,
    filecolor=magenta,      
    urlcolor=cyan,
    citecolor=blue
}
\usepackage{cleveref}

\begin{document}

\title{\bf In-Context Clustering with Large Language Models} 
\author{
  Ying Wang, Mengye Ren, Andrew Gordon Wilson \\
  New York University\\
  {\small \texttt{\{yw3076, mengye, aw130\}@nyu.edu}}
}
\date{}
\maketitle

\begin{abstract}
We propose \emph{In-Context Clustering} (ICC), a flexible LLM-based procedure for clustering data from diverse distributions. Unlike traditional clustering algorithms constrained by predefined similarity measures, ICC flexibly captures complex relationships among inputs through an attention mechanism. We show that pretrained LLMs exhibit impressive zero-shot clustering capabilities on text-encoded numeric data, with attention matrices showing salient cluster patterns. Spectral clustering using attention matrices offers surprisingly competitive performance. We further enhance the clustering capabilities of LLMs on numeric and image data through fine-tuning using the Next Token Prediction (NTP) loss. Moreover, the flexibility of LLM prompting enables text-conditioned image clustering, a capability that classical clustering methods lack. Our work extends in-context learning to an unsupervised setting, showcasing the effectiveness and flexibility of LLMs for clustering. Our code is available at \url{https://agenticlearning.ai/icc}.
\end{abstract}    
\section{Introduction}
Central to any clustering procedure is a similarity measure that makes it possible to separate data into meaningful groups. Classical methods often rely on predefined measures, such as k-means with Euclidean distance, and therefore impose strong assumptions on the underlying data distributions. As a result, these approaches often struggle with high-dimensional and semantically complex data such as text~\citep{text_clustering, shah2012document}, images~\citep{image_analysis_clustering_survey_WAZARKAR2018596, adaptive_image_clustering_Chang_2017_ICCV, guerin2018improving_image_cnn}, and audio~\citep{audio_clustering_Meinedo2023, NEURIPS2020_audio_video_clustering}, where similarity is context-dependent and cannot be easily captured by a rigid predefined function. 

Recent advances in Large Language Models (LLMs) offer a promising alternative through in-context learning (ICL)~\citep{NIPS2017_attention_all_you_need, NEURIPS2020_fewshotlearner}, which has been proven effective across a variety of data distributions~\citep{tsimpoukelli2021multimodal_fewshot, garg2022llm_simple_function, gruver2023llmtime, vacareanu2024words}. Instead of using a predefined similarity function, LLMs capture context-dependent relations through an attention mechanism with query and key projections learned from large-scale pretraining. The ability to recognize contextual relationships among in-context examples provides a foundation for flexible clustering that can adapt to diverse data and different criteria. This LLM-based approach particularly excels in \emph{few-shot scenarios involving semantically rich, naturalistic data}, complementing classical methods optimized for structured large-scale datasets.

\begin{figure*}[t]
\center
\includegraphics[width=\textwidth]{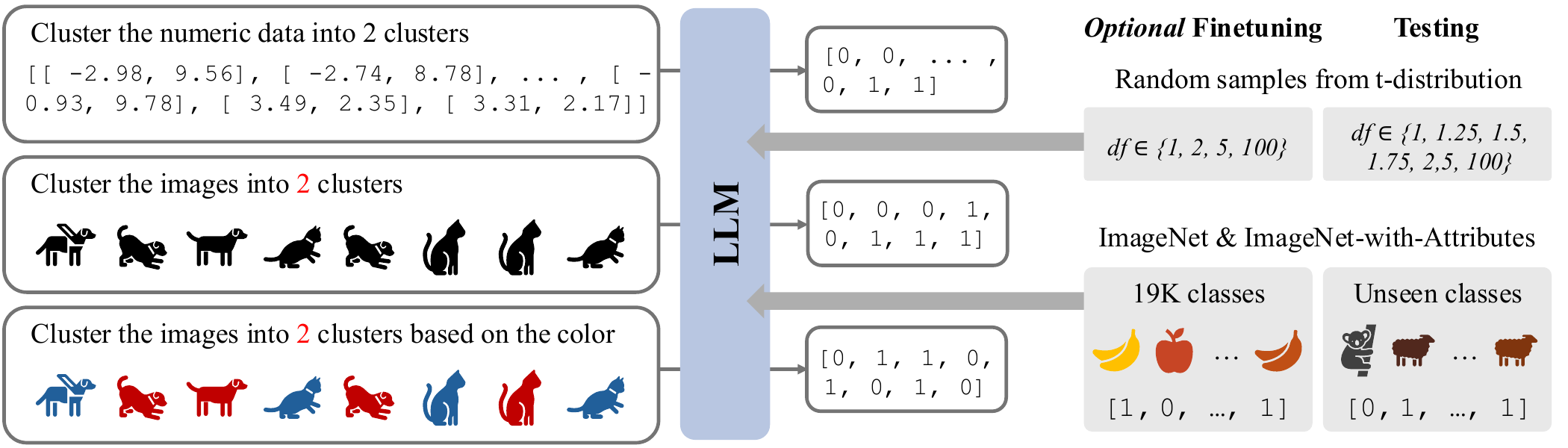}
\caption{\emph{In-Context Clustering} (ICC). LLMs can flexibly handle diverse modalities and perform text-conditioned clustering. We show the zero-shot clustering capability in pretrained LLMs and further strengthen it through finetuning.
}
\label{img:teaser}
\vspace{-0.1in}
\end{figure*}

In this work, we propose \emph{In-Context Clustering}~(ICC), extending in-context learning to an unsupervised setting (\Cref{img:teaser}). Different from previous in-context supervised learning that requires multiple input-output pairs in the prompt~\citep{NEURIPS2020_fewshotlearner}, ICC utilizes only unlabeled input data in the context. Given a natural language instruction specifying the clustering objective and a sequence of inputs, the LLM generates cluster labels autoregressively. When the clustering condition changes (e.g., grouping by color instead of class as shown in~\Cref{img:img_cond}), one can simply modify the prompt without updating model weights or features. We evaluate ICC on numerical data and image data using a variety of synthetic and real-world datasets to demonstrate the effectiveness and flexibility of ICC. 

Our paper is structured as follows:
\begin{itemize}[leftmargin=*]
    \item We demonstrate that LLMs can provide surprisingly strong zero-shot in-context clustering capabilities (\Cref{sec:zero_shot_prompt}).  
    \item We find attention matrices in intermediate layers show salient cluster structures. Moreover, spectral clustering using these attention matrices yields impressive performance (\Cref{sec:attention}).
    \item With lightweight LoRA fine-tuning~\citep{hu2021lora} using NTP loss on generated clustering data, we find ICC significantly improves on numeric (\Cref{sec:finetune_num}) and image data (\Cref{sec:finetune_image}), especially under heavy-tailed distributions and for images with rich semantics.
    \item 
    We show that ICC has the relatively distinct ability to do text-conditional image clustering, demonstrating flexibility beyond classical methods. For example, ``cluster based on color'', or ``cluster based on foreground''. 
    We believe that this ability to change the way clustering is done based on different prompts makes ICC, and this research direction, particularly compelling. Finally, we show ICC outperforms recent caption-based LLM clustering~\citep{kwon2024imageclusteringtext} (\Cref{sec:conditonal}).
\end{itemize}

\section{Related Work}

\paragraph{Classical Clustering Algorithms.} 
Classical clustering methods can be classified into hierarchical, partitional, and density-based methods~\citep{traditional_clustering_survey_1999, image_analysis_clustering_survey_WAZARKAR2018596}. Hierarchical methods continuously merge data points into clusters based on their similarity with others, resulting in a dendrogram of the data~\citep{ward1963hierarchical, murtagh2012hierarchical_clustering}. By contrast, partitional clustering algorithms output a single partition of the data instead of a clustering hierarchy~\citep{kmeans-review}. K-means is one of the most widely used partitional clustering methods based on Euclidean distance and works well for spherical Gaussian clusters. Density-based methods can find arbitrarily shaped clusters by detecting the dense regions in the given dataset~\citep{dbscan}. Although widely used, classical methods lack the ability to do representation learning, instead relying on predefined similarity measures that make strong or often unrealistic assumptions about the data. These drawbacks motivate a more flexible clustering algorithm effective for diverse distributions.

\paragraph{LLMs for Text Clustering.} LLMs have demonstrated their excellent ability to understand and reason with natural language~\citep{sparks, huang-chang-2023-llm-survey, zhang2024llmmastermindsurveystrategic}. Recent studies have demonstrated the effectiveness of LLMs in text clustering~\citep{zhang-etal-2023-clusterllm,viswanathan-etal-2024-fewshot-text-clustering, nakshatri-etal-2023-using, tipirneni2024contextawareclusteringusinglarge}.  Various strategies have been explored to enhance clustering performance, including LLM-generated embeddings~\citep{zhang-etal-2023-clusterllm} and few-shot prompting~\citep{viswanathan-etal-2024-fewshot-text-clustering}. However, these practices are limited to text, where the success is somewhat expected, given that the input aligns closely with the pre-training data of the LLMs. In this paper, we extend LLM clustering to non-textual modalities. We find that language pretaining provides a strong foundation for clustering numeric and imagery data.

\paragraph{Multimodal Clustering.} \looseness=-100000 Multimodal data introduces challenges in aligning heterogeneous information across modalities. Clustering can be performed jointly across modalities using a shared embedding space, or conditionally where one modality guides the clustering of another. As an example for joint multimodal clustering,~\citet{su2024multimodalgeneralizedcategorydiscovery} propose Multimodal Generalized Category Discovery (Multimodal GCD) that focuses on partitioning a shared multimodal embedding space into known and novel categories. As for conditional multimodal clustering, \ictc{}~\citep{kwon2024imageclusteringtext} and SSD-LLM~\citep{luo2025ssdllm} both leverage LLMs for text-conditioned image clustering by converting images to captions. \ictc{} distills image captions into one-word labels using an LLM, which are clustered according to the given textual criteria, and the final assignment is made by prompting the LLM to match image captions to the cluster labels. SSD-LLM uses LLMs iteratively to refine and produce subpopulation structures based on image captions, and then utilizes the subpopulation structures for clustering. While the task of text-conditioned image clustering is similar to ours in~\Cref{sec:conditonal}, these caption-based approaches are highly constrained by the caption quality, failing to generalize when the data has complicated or nuanced relationships that the captioner is unable to capture.

\begin{figure}[t]
\centering
\includegraphics[width=\textwidth]{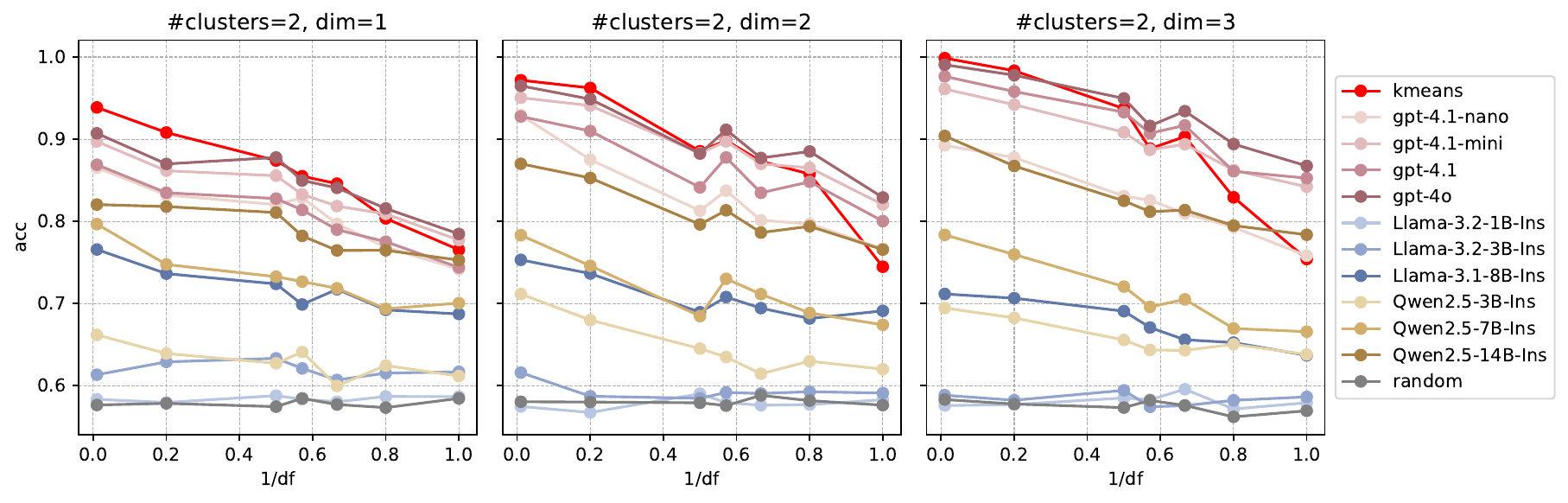}
\caption{\looseness=-100000 Zero-shot Clustering Accuracy on $t$-Distribution with Different Degrees of Freedom. When $df$ is small, the data distribution has a heavy tail, which violates the Gaussian assumption of k-means. LLMs show impressive zero-shot clustering capabilities on heavy-tailed data. 
}
\label{img:zeroshot}
\end{figure}

\section{Zero-shot Clustering}
\label{sec:zeroshot}

In this section, we show that LLMs pre-trained on large text corpus are capable of zero-shot clustering. LLMs outperform k-means on non-Gaussian data, demonstrating their potential to perform in-context clustering. We also observe that a cluster-like pattern emerges in the self-attention of pretrained LLMs and using the attention matrices for spectral clustering results in competitive performance.

\subsection{Zero-shot In-Context Clustering}
\label{sec:zero_shot_prompt}

\paragraph{Experimental Setup.} To understand the zero-shot clustering capabilities of different model families and model sizes, we test pre-trained Llama 3.1\&3.2~\citep{llama3}, Qwen 2.5~\citep{qwen} with different sizes, and various closed-source GPT models~\citep{gpt4} including \textsc{GPT-4o} and \textsc{GPT-4.1} series. We round all numbers to two decimal places and use text to represent the input numeric data as a double list where the inner list represents one data point. Our prompt is as follows:

\begin{quote}
\emph{Cluster the following data into \textcolor{blue}{\{\#clusters\}} clusters. Only output the cluster labels for each point as a list of integers. Data: \textcolor{blue}{\{input data\}} Labels:}
\end{quote}

\paragraph{Data.} We sample data from a $t$-distribution to evaluate ICC under diverse conditions: When $df$ are large, it approximates the Gaussian distribution; when $df$ are small, it exhibits a heavy tail. We first sample the cluster centroids by drawing each dimension uniformly from $[-10, 10]$, and then generate data points within each cluster by sampling from a $t$-distribution with the specified $df$. For each  combination of the number of clusters $c\in\{2,3,4,5\}$,  dimensions $d\in\{1,2,3,4\}$, and different degrees of freedom $df \in \{1,1.25,1.5,1.75,2,5,100\}$, we generate 100 samples with length randomly drawn from $[10,50]$. The size of each cluster is also random but forced to be nonempty.

\paragraph{Results.} We report zero-shot accuracy\footnote{Since clustering is invariant to label permutation, we adopt the Hungarian Algorithm to find the optimal assignment before computing the accuracy.} in~\Cref{img:zeroshot} and include more results of different numbers of clusters and dimensions in~\Cref{img:zeroshot_full} of~\Cref{sec:appendix_numeric}. LLMs show impressive zero-shot clustering capabilities, outperforming k-means when the data has heavy tails. When $df$ is small, the Gaussian assumption of k-means is violated, leading to a significant drop in performance. \textsc{gpt-4} and \textsc{gpt-4.1} outperform k-means when data is heavy-tailed and high-dimensional, demonstrating the potential of applying LLMs for clustering high-dimensional non-Gaussian data.

The performance of LLMs is correlated with the model size and training choices. Small LLMs with 3B or 8B parameters can produce non-trivial answers when the clustering data is simple (with lower dimensions and fewer clusters, shown in~\Cref{img:zeroshot_full}). When the data becomes more complicated, these small LLMs are either unable to follow the instruction of generating the correct number of clusters or produce answers that are close to random guesses. We also observe that instruction tuning improves the overall accuracy, without which the model is unable to follow the instructions of the clustering task (\Cref{img:finetune_instruct}). There is still a gap between the performance of small open-source models and GPT models, probably due to the difference in the model size and pretraining. In~\Cref{sec:finetune}, we show that finetuning Llama models on synthetic clustering data helps close the gap. 

\subsection{Emergence of Clusters in Attention}
\label{sec:attention}

\begin{figure}[t]
\centering
\vspace{-0.05in}
\includegraphics[width=\textwidth]{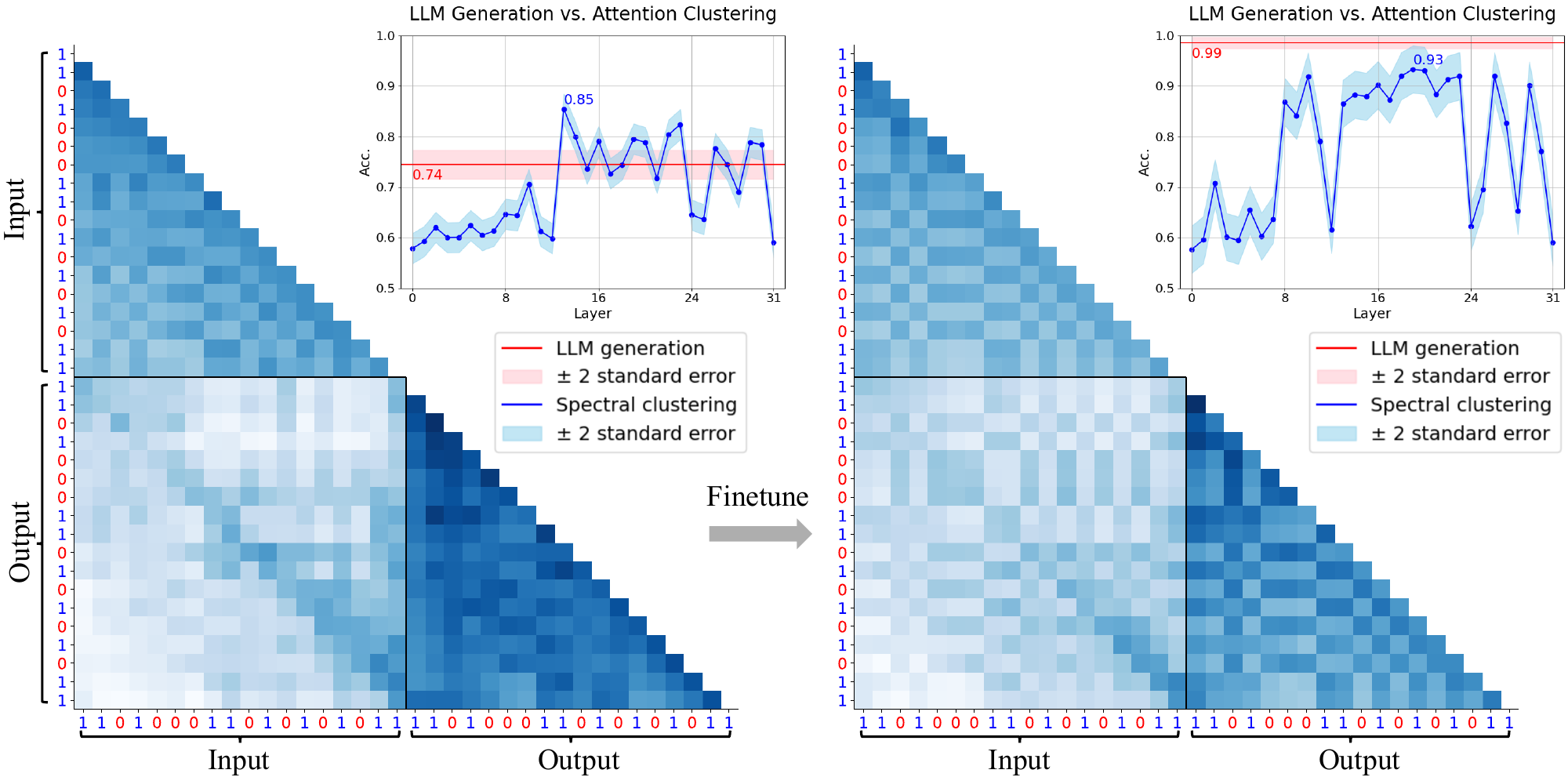}
\caption{Visualization of Attention Allocation of Input Data and Generated Cluster Labels at an Intermediate Layer. The x-axis and y-axis are the ground-truth cluster labels. The left figure is for the pretrained \textsc{Llama-3.1-8b-Instruct}, and the right is after fine-tuning(details in~\Cref{sec:finetune_num}). The top right curves are the average accuracy of spectral clustering using the input-input attention score matrices (top-left) across different layers, compared with the average accuracy of LLM generation. 
}
\vspace{-0.1in}
\label{img:attention}
\end{figure}

To better understand the inner mechanism of ICC, we visualize the attention scores across different transformer layers. All LLMs considered here are causal transformers with multi-head self-attention. Given a textual prompt as described in~\Cref{sec:zeroshot}, the model autoregressively generates cluster labels conditioned on the input data and previous generation. At each layer, we extract the self-attention matrix $A \in \mathbb{R}^{n \times n}$, a lower-triangular matrix due to causality, where $n$ is the total number of tokens. For multi-head attention, we use average attention scores across heads in this section.

To focus on input data and output cluster label tokens, we discard instruction and system prompt tokens. Since each input data point may span multiple tokens, we aggregate token-level attention scores to obtain data-level attention scores. Let $m$ denote the number of input data points. From the full matrix $A$, we construct an aggregated attention matrix with the following block structure: 
\begin{equation}
A = 
\begin{bmatrix}
A^{II} & 0 \\
A^{OI} & A^{OO}
\end{bmatrix}.
\end{equation}
Here, $A^{II} \in \mathbb{R}^{m \times m}$ represents the input-input matrix capturing attention scores among input data points, $A^{OI} \in \mathbb{R}^{m \times m}$ represents the output-input matrix that reflects how generated cluster labels attend to input data, and $A^{OO} \in \mathbb{R}^{m \times m}$ represents the output-output matrix containing attention scores among output tokens. Each input data point $d_i$ may span multiple tokens, indexed from $s_i$ to $e_i$. We compute $A^{II}$ by averaging attention scores across all token pairs between $d_i$ and $d_j$:
\begin{equation}
    A^{II}_{ij} := \frac{1}{(e_i - s_i + 1)(e_j - s_j + 1)} \sum_{p=s_i}^{e_i} \sum_{q=s_j}^{e_j} A_{pq}.
\end{equation}
Each output cluster label is represented by a single token, indexed as $t_i$ for the label of $d_i$. The remaining attention blocks are defined as:
\begin{equation}
A^{OI}_{ij} := \frac{1}{e_j - s_j + 1} \sum_{p = s_j}^{e_j} A_{t_ip}, \;\;\; \;
A^{OO}_{ij} := A_{t_i t_j}.
\end{equation}
\vspace{-0.01in}\Cref{img:attention} visualizes this block matrix , with $A^{II}$ in the top-left, $A^{OI}$ in the bottom-left, and $A^{OO}$ in the bottom-right. Here, we take one clustering example generated from Gaussian distribution with two clusters. We observe that \emph{attention matrices in intermediate layers show block structures that align with cluster identities}. The transformer assigns higher attention scores to similar data within the same cluster that has been seen in the past. We provide more examples across different layers in~\Cref{sec:appendix_attention}. This cluster pattern is consistent and salient in most middle layers. In contrast, the final layer typically shows a vertical-slash pattern, as also observed by~\citet{jiang2024minference}. We also observe that most attention heads show similar cluster patterns in~\Cref{img:attention_heads}.

Although the pretrained model (left in \Cref{img:attention}) has a clear cluster pattern in the input-input matrix, clusters are not observed in attention related to outputs. This suggests that the model learns similarity among input data during pretraining, but is not optimized for generating cluster labels as explicit clustering tasks are very likely rare in pretraining.\footnote{Llama 3 models are claimed to be trained on "15T tokens that were all collected from publicly available sources"\citep{llama3}, but details are not disclosed.} After fine-tuning on ICC data, the cluster structure in the input-input matrix becomes stronger, and similar clusters also emerge in output-input and output-output matrices. 

To quantify how well the attention captures the similarity among the input data, we use these input-input attention score matrices for spectral clustering~\citep{NIPS2001_spectral_clustering, vonluxburg2007tutorialspectralclustering} (more details and results are in~\Cref{sec:appendix_spectral}). Although the zero-shot accuracy of prompting pretrained \textsc{Llama-3.1-8b-Instruct} to cluster is 74\%, the spectral clustering using attention with the optimal choice of layers achieves 85\% before fine-tuning. This surprising result suggests that attention of LLMs already encodes rich structural information beyond what is directly generated. In addition to prompting the LLM for generation, directly using attention can be an alternative to leverage pretrained LLM for in-context clustering in zero shot.
\vspace{-0.05in}

\section{Learning Clustering with Next Token Prediction}
\label{sec:finetune}

While pretrained LLMs show promising zero-shot clustering capabilities, small open-source models lag behind classical methods and proprietary LLMs. In this section, we show that the clusterng capabilities of pretrained LLMs can be further enhanced through LoRA fine-tuning using NTP loss. Inspired by the meta learning literature~\citep{ ravi2017optimizationfewshot, min-etal-2022-metaicl,najdenkoska2023metavlm}, we construct various clustering episodes to make pretrained (multimodal) LLM learn to cluster in context and then test it on unseen classes. We experiment on both numeric and image data.

\vspace{-0.05in}
\subsection{Numeric Data Clustering}
\label{sec:finetune_num}

\paragraph{Experiment Setup.} We follow the standard Supervised Fine-Tuning (SFT) procedure to fine-tune pre-trained Llama models with different sizes (\textsc{Llama-3.2-1B-Instruct}, \textsc{Llama-3.2-3B-Instruct}, \textsc{Llama-3.1-8B-Instruct}) using NTP loss. Similarly to how we construct the clustering data in~\Cref{sec:zeroshot}, we construct the data by randomly sampling data from a $t$-distribution with different degrees of freedom $df\in\{1,2,5,100\}$, the number of clusters $c\in\{2,3,4,5\}$, and dimensions of each point $d\in\{1,2,3,4\}$. We generate around 100k input-label pairs, where each sample has a length randomly drawn from $[10,50]$. We use LoRA~\citep{hu2021lora} to fine-tune the pre-trained Llama model for one epoch with an effective batch size of 32 and a learning rate of 5e-4.

\begin{table}[t]
\caption{Effect of Finetuning on $t$-Distributed Data with Different Degrees of Freedom. Input $dim=3$ and number of clusters $c=3$. We report average accuracy (\%) and one standard error.
}
\vspace{-0.1in}
\label{tbl:numeric_finetune}
\begin{center}
\begin{small}
\begin{sc}
\resizebox{\columnwidth}{!}{
\begin{tabular}{l|ccccccc}
\toprule
     & \cellcolor{Gainsboro!60}  df=1       &  df=1.25 &  df=1.5       & df=1.75       & \cellcolor{Gainsboro!60} df=2      & \cellcolor{Gainsboro!60} df=5       & \cellcolor{Gainsboro!60} df=100       \\
\midrule
kmeans & 67.95\tiny{$\pm$1.46} & 75.43\tiny{$\pm$1.52} & 85.57\tiny{$\pm$1.20} & 87.55\tiny{$\pm$1.32} & 89.05\tiny{$\pm$1.27} & 95.29\tiny{$\pm$1.00} & 97.08\tiny{$\pm$0.82} \\

gpt-4o & 77.75\tiny{$\pm$1.31} & 80.60\tiny{$\pm$1.20} & 86.99\tiny{$\pm$1.15} & 87.08\tiny{$\pm$1.26} & 89.56\tiny{$\pm$1.10} & 93.84\tiny{$\pm$1.03} & 96.25\tiny{$\pm$0.86} \\

(a) Llama-3.2-1B-Instruct & 45.40\tiny{$\pm$0.64} & 47.09\tiny{$\pm$0.71} & 46.77\tiny{$\pm$0.66} & 46.63\tiny{$\pm$0.67} & 46.54\tiny{$\pm$0.69} & 45.73\tiny{$\pm$0.64} & 47.36\tiny{$\pm$0.77} \\
\rowcolor{Blue!10}
(a) + finetune & 82.66\tiny{$\pm$1.30} & 86.45\tiny{$\pm$1.23} & 91.10\tiny{$\pm$0.90} & 89.46\tiny{$\pm$1.18} & 88.76\tiny{$\pm$1.20} & 95.09\tiny{$\pm$0.93} & 96.28\tiny{$\pm$0.88} \\

(b) Llama-3.2-3B-Instruct & 46.71\tiny{$\pm$0.67} & 46.09\tiny{$\pm$0.72} & 46.35\tiny{$\pm$0.62} & 46.85\tiny{$\pm$0.76} & 46.05\tiny{$\pm$0.82} & 46.84\tiny{$\pm$0.72} & 46.35\tiny{$\pm$0.86} \\
\rowcolor{Blue!10}
(b) + finetune & 88.54\tiny{$\pm$1.03} & 91.05\tiny{$\pm$1.00} & 94.31\tiny{$\pm$0.77} & 93.33\tiny{$\pm$0.90} & 94.51\tiny{$\pm$0.90} & 98.08\tiny{$\pm$0.49} & 97.64\tiny{$\pm$0.78} \\

(c) Llama-3.1-8B-Instruct & 55.29\tiny{$\pm$1.34} & 55.38\tiny{$\pm$1.44} & 59.80\tiny{$\pm$1.57} & 61.09\tiny{$\pm$1.55} & 61.21\tiny{$\pm$1.47} & 64.73\tiny{$\pm$1.66} & 64.42\tiny{$\pm$1.73} \\
\rowcolor{Blue!10}
(c) + finetune & \textbf{90.66}\tiny{$\pm$0.95} & \textbf{92.20}\tiny{$\pm$0.93} & \textbf{95.25}\tiny{$\pm$0.54} & \textbf{94.57}\tiny{$\pm$0.86} & \textbf{95.44}\tiny{$\pm$0.71} & \textbf{98.90}\tiny{$\pm$0.31} & \textbf{97.85}\tiny{$\pm$0.76} \\
\bottomrule
\end{tabular}
}
\vspace{-0.2in}
\end{sc}
\end{small}
\end{center}
\end{table}

\paragraph{Results.} We use the test data in~\Cref{sec:zeroshot} ($df \in \{1,1.25,1.5,1.75,2,5,100\}$) with $df \in \{1.25,1.5,1.75\}$ to test the robustness of the fine-tuned model. During fine-tuning, the LLM exhibits a two-phase learning pattern where it first learns the correct format and then gradually develops a clustering mechanism. Initially, the LLM (especially smaller models with 1B or 3B parameters) struggles with instruction following and produces repetitive outputs. These poorly formatted predictions are heavily penalized by the NTP loss. As training progresses, the model learns to effectively differentiate among cluster labels based on the input data and achieves a high accuracy.

\looseness=-100000
As shown in~\Cref{tbl:numeric_finetune}, all fine-tuned models show superior performance compared to k-means and \textsc{gpt-4o} (the complete results are in~\Cref{img:finetune_instruct} of~\Cref{sec:appendix_numeric}). Although these LLMs are fine-tuned on $t$-distributed data with $df\in\{1,2,5,100\}$, they show generalization capability to more $df$ and different distributions. All fine-tuned models perform consistently well on $t$-distributed data with new~$df\in\{1.25,1.5,1.75\}$. While these models are fine-tuned on a symmetric distribution, they also significantly outperform k-means and \textsc{gpt-4o} on a skewed distribution (lognormal) as shown in~\Cref{tbl:lognormal} in~\Cref{sec:appendix_numeric}. We also observe that models with higher accuracy tend to be more invariant to
permutation in input data, and data augmentation is effective in improving consistency, as shown in~\Cref{tbl:permutation_invariance}.

We study the effect of fine-tuning by analyzing the attention pattern as visualized in~\Cref{img:attention}. The cluster pattern in the attention score matrix of the input data is significantly more salient after fine-tuning, indicating that the model learns a better similarity function among the data through its attention mechanism during fine-tuning. The accuracy of spectral clustering using attention scores increases as well. More visualization and results are in~\Cref{sec:appendix_attention_clustering}.

\subsection{Image Clustering} 
\label{sec:finetune_image}

Here, we extend ICC to multimodal LLMs and present results of image clustering. Given a set of images, the goal is to cluster based on their semantic meanings. By projecting image embeddings obtained from a pretrained visual encoder, LLMs can learn to produce meaningful groupings that outperform an LLM-based method that relies on image captions.

\paragraph{Model.} We use \texttt{llava-interleave-qwen-7b-hf}~\citep{li2024llavanextinterleavetacklingmultiimagevideo}, a multimodal LLM pretrained with multi-image inputs, as our base model. In the LLaVA framework, each image is segmented into 729 patches encoded by a pre-trained ViT, namely the SigLIP's visual encoder~\citep{zhai2023siglip}, then projected through an MLP layer into the embedding space of the base LLM~\citep{qwen}. While such a high-granularity representation may benefit downstream tasks like object detection, we argue that it is not optimal for clustering tasks. Clustering typically involves a large number of images; thus, using hundreds of tokens per image can quickly exceed context length limitations and significantly increase computational costs during fine-tuning. Additionally, high granularity might be unnecessary for some clustering tasks that only rely on global features. 

\begin{figure*}[t]
\center
\includegraphics[width=\textwidth]{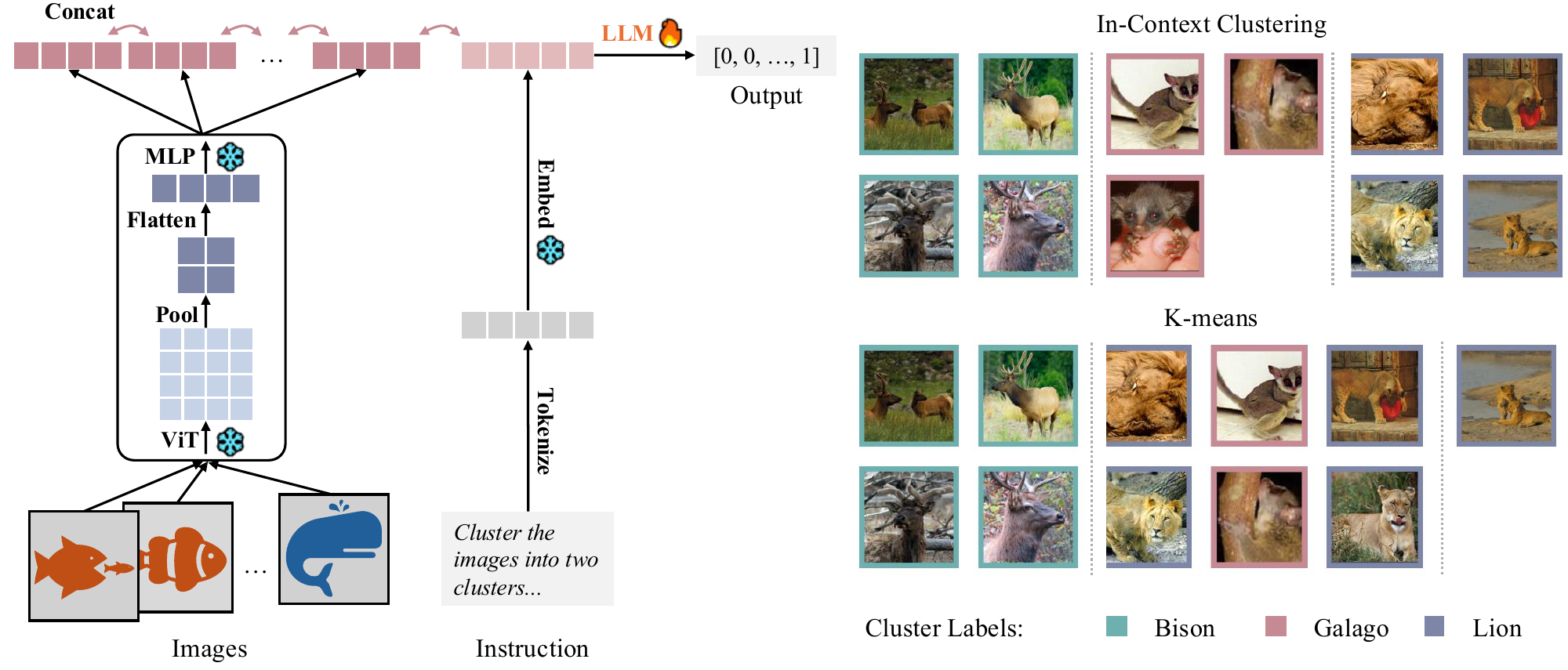}
\caption{Left: Multimodal LLM Architecture with Average Pooling for Image Features. Right: Qualitative Comparison of Models on Image Clustering --- ICC outperforms k-means when the data has rich semantic information. 
}
\label{img:img_clustering}
\end{figure*}

To address these efficiency concerns, we implement average pooling after the projection layer to reduce per-image token lengths, as illustrated in~\Cref{img:img_clustering}~(left). Each input image is divided into patches, which are preprocessed and flattened (omitted from the figure for clarity), and then encoded by a vision transformer. We reshape the flattened image features back to 2D and then apply average pooling to reduce dimensionality. The pooled features are then flattened, projected into the LLM's embedding space, and concatenated with text token embeddings. We experiment with various pooling kernel sizes in~\Cref{sec:appendix_pooling}. No padding is applied and the stride is the same as the kernel width.

\paragraph{Data.} We collect images from ImageNet21k~\citep{ridnik2021imagenet21k} where images sharing the same label are considered part of the same cluster. We reserve the 384 image classes covered in ImageNet-with-Attributes~\citep{RussakovskyECCV10_imagenet_attr} for testing and the remaining 18K classes for training. For training, we construct 192K image clustering episodes of various numbers of clusters $c \in \{2,3,4\}$, with random length $l \in [10,30]$ and random cluster proportion. For testing, we use the reserved test classes to construct 100 clustering episodes for each number of clusters. To test generalization on out-of-domain data, we include Plant Disease and EuroSAT datasets from the Cross-Domain Few-Shot Learning (CD-FSL) Benchmark~\citep{guo2020cdfsl} with details in~\Cref{sec:cdfsl}.

\paragraph{Experiment Setup.} Similarly to previous numerical experiments, we use LoRA to fine-tune the LLM with NTP loss. The visual encoder and projection layer are frozen during training. We fine-tune for one epoch with an effective batch size of 32 and a learning rate of 5e-4. 

\paragraph{Baselines.} To ensure a fair comparison, we use average-pooled image features from the vision encoder of the base model~\citep{li2024llavanextinterleavetacklingmultiimagevideo} as the inputs to k-means. We also compare ICC against \ictc~\citep{kwon2024imageclusteringtext}, a recent LLM-based image clustering method. We use the same model~\citep{li2024llavanextinterleavetacklingmultiimagevideo} to generate image captions for \ictc{} then use \textsc{gpt-3.5-turbo} to distill and cluster the captions according to the given number of clusters and the clustering condition. Although converting images to short captions facilitates clustering via LLMs, \ictc{} experiences information loss during the captioning and summarization stage, limiting its performance on challenging data.

\begin{table*}[t]
\caption{Image Clustering Accuracy (\%) with Standard Error. \textsc{ICC(gpt-4o)} is zero-shot ICC using gpt-4o and the shaded rows represent models finetuned on ImageNet data with numbers of clusters $c \in \{2,3,4\}$, where \textsc{Small, Medium, Large} refer to the per-image token length in~\Cref{sec:appendix_pooling}. Our finetuned models can generalize to unseen $c=5$ and other datasets that deviate from ImageNet. }
\vspace{-0.1in}
\label{tbl:img_results}
\begin{center}
\begin{small}
\begin{sc}
\resizebox{0.85\textwidth}{!}{
\begin{tabular}{lcccccc}
\toprule
                       & \multicolumn{4}{c}{ImageNet} & Plant & EuroSAT \\
number of clusters     &  c=2       &  c=3       &  c=4       & c=5  &  c=2 &  c=2   \\
\midrule
k-means           & 89.43{\tiny$\pm$1.57} & 82.09{\tiny$\pm$1.44} & 79.07{\tiny$\pm$1.31} & 77.96{\tiny$\pm$1.08} & \textbf{93.70}{\tiny$\pm$1.40} & \textbf{85.52}{\tiny$\pm$1.43} \\
\ictc \citep{kwon2024imageclusteringtext} & 90.20{\tiny$\pm$1.54} & 78.86{\tiny$\pm$1.41} & 76.49{\tiny$\pm$1.50} & 73.99{\tiny$\pm$1.58} & 67.40{\tiny$\pm$1.23} & 72.97{\tiny$\pm$1.42} \\
ICC (gpt-4o)   & 82.46{\tiny$\pm$1.40} & 80.25{\tiny$\pm$1.73} & 75.91{\tiny$\pm$1.73} & 78.08{\tiny$\pm$1.50} & 84.74{\tiny$\pm$1.25} & \underline{79.08}{\tiny$\pm$1.41} \\
\rowcolor{Blue!10}
ICC (Small)   & 96.81{\tiny$\pm$0.83} & 91.94{\tiny$\pm$1.03} & 89.83{\tiny$\pm$1.19} & 82.08{\tiny$\pm$1.01} & 73.03{\tiny$\pm$1.58} & 78.17{\tiny$\pm$1.53} \\
\rowcolor{Blue!10}
ICC (Medium)   & \underline{98.26}{\tiny$\pm$0.71} & \textbf{95.92}{\tiny$\pm$0.90} & \underline{91.62}{\tiny$\pm$1.16} & \underline{84.92}{\tiny$\pm$0.95} & 82.28{\tiny$\pm$1.85} & 78.64{\tiny$\pm$1.61} \\
\rowcolor{Blue!10}
ICC (Large)    & \textbf{99.12}{\tiny$\pm$0.41} & \underline{91.95}{\tiny$\pm$0.96} & \textbf{92.92}{\tiny$\pm$1.06} & \textbf{84.96}{\tiny$\pm$0.89} & \underline{85.09}{\tiny$\pm$1.80} & 77.35{\tiny$\pm$1.70} \\
\bottomrule
\end{tabular}
}
\vspace{-0.2in}
\end{sc}
\end{small}
\end{center}
\end{table*}

\paragraph{Results.} The performance of different models is summarized in~\Cref{tbl:img_results}. While zero-shot ICC using \textsc{gpt-4o} achieves competitive performance, it is less effective than on text-encoded data. This is likely due to the current limitations of multimodal LLMs on long sequences of complex images. Our proposed finetuning method significantly closes this gap, achieving strong performance across all datasets. Despite being only fine-tuned on ImageNet data with the number of clusters less than five, our model can generalize to within-domain data of five clusters and out-of-domain data including plant leaves and satellite images. 

With good image features, k-means is effective on datasets with limited semantic complexity, such as Plant Disease and EuroSAT. However, it loses its competence on ImageNet, where images often depict complex scenes involving multiple objects. The caption-based method, \ictc, performs poorly on Plant Disease or EuroSAT, as its captioning model lacks domain-specific knowledge. This observation highlights a key weakness of caption-based clustering: its dependence on accurate and relevant captions limits its applicability to novel domains. Our model avoids these pitfalls, demonstrating superior flexibility and performance across both general and specialized domains.

\section{Text-Conditioned Clustering}
\label{sec:conditonal}

\begin{figure}[t]
\centering
\includegraphics[width=\textwidth]{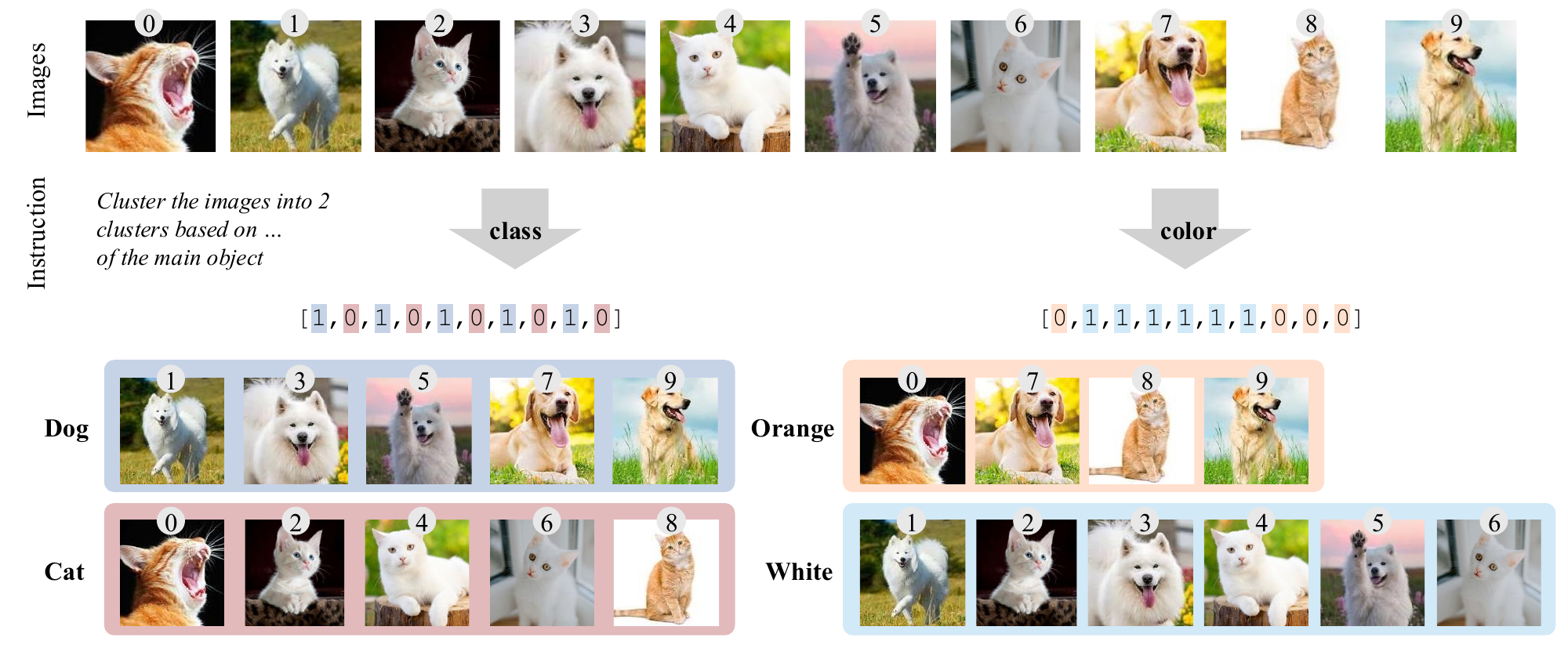}
\caption{LLMs are able to produce different clusterings according to the condition in the prompt.
}
\label{img:img_cond}
\end{figure}

While the experiments in the previous section assume a single, fixed clustering objective, real-world data admits multiple plausible clusterings depending on the objective. For example, the same set of animal images can be clustered by visual properties like colors (orange vs. white) or semantic categories like species (dog vs. cat), as shown in~\Cref{img:img_cond}. When the clustering condition changes, classical methods typically require retraining or re-engineering features. In contrast, LLMs can easily adapt to new conditions through prompting thanks to their powerful contextual understanding capability. In this section, we perform text-conditioned image clustering by fine-tuning multimodal LLMs with the NTP loss. 

\paragraph{Data.} We construct conditional clustering using ImageNet-with-Attributes~\citep{RussakovskyECCV10_imagenet_attr}, which includes 384 classes with 4 categories of attributes (\textsc{color}, \textsc{shape}, \textsc{pattern}, \textsc{texture}). We split the data into 80\% training classes and 20\% testing classes. We treat the category name as the clustering condition that will be specified in the prompt and use the attribute value as cluster labels. In addition, we include an \textsc{object} category that is similar to~\Cref{sec:finetune_image}, where we use the class name of the images as cluster labels. Images with ambiguous annotations are filtered out. For training, we construct around 280K image conditional clustering episodes of various numbers of clusters $c \in \{2,3,4\}$,\footnote{The pattern category only has two available values, so we don't have $c \in \{2,3\}$ for this category.} with random length $l \in [10,30]$ and random cluster proportion. 

To test the performance of the model on different conditions, we use the reserved test classes of ImageNet-with-Attributes and also include the Stanford 40 Action dataset~\citep{stanford_action} with annotations on the \textsc{location} of the scene, the \textsc{action} and \textsc{mood} of the people in the image provided by~\citep{kwon2024imageclusteringtext}. For each dataset and clustering condition, we sample 100 clustering data from two random classes of each attribute category, with random size $l \in [10,30]$ and random cluster proportion. 

\paragraph{Experiment Setup.} Following the SFT procedure in~\Cref{sec:finetune_image}, we use LoRA to fine-tune \\ \texttt{llava-interleave-qwen-7b-hf} with different pooling ratios. We keep the visual encoder and projection layer frozen during training. We use NTP loss to fine-tune for one epoch with an effective batch size of 32 and a learning rate of 5e-4. 

\paragraph{Baselines.} We test both unconditional and conditional clustering methods. K-means is a unconditional baseline as it does not allow injecting clustering criteria. For conditional clustering methods, we test \ictc{} explicitly specifying conditions in the prompts for all the summarization and clustering stages, with \textsc{gpt-3.5-turbo} as the LLM to save costs.

\begin{table}[t]
\caption{Conditional Image Clustering Accuracy (\%) with Standard Error. Here, \textsc{ICC (Medium:\ref{sec:finetune_image})} represents the model finetuned on unconditional image clustering data in~\Cref{sec:finetune_image}, while others use conditional image clustering data in~\Cref{sec:conditonal}. Our method outperforms all baselines on ImageNet and Stanford 40 Action. \textsc{Small, Median, Large} refer to the per-image token length in~\Cref{sec:appendix_pooling}. 
}
\label{tbl:img_cond}
\begin{center}
\begin{small}
\begin{sc}
\resizebox{\textwidth}{!}{
\begin{tabular}{lcccccccc}
\toprule
                       & \multicolumn{5}{c}{ImageNet} & \multicolumn{3}{c}{Stanford 40 Action} \\
                       &  object       &  color       &  pattern   & shape   & texture & action & mood & location   \\
\midrule
\multicolumn{9}{l}{\textit{Unconditional Methods}} \\
k-means           & 89.96{\tiny$\pm$1.44} & 66.40{\tiny$\pm$1.16} & 62.36{\tiny$\pm$0.98} & 75.76{\tiny$\pm$1.78} & 78.53{\tiny$\pm$1.65} & 79.90{\tiny$\pm$1.76} & 70.93{\tiny$\pm$1.43} & 78.11{\tiny$\pm$1.50} \\
\midrule
\multicolumn{9}{l}{\textit{Conditional Methods}} \\
\ictc \citep{kwon2024imageclusteringtext}  & 91.93{\tiny$\pm$1.38} & 69.70{\tiny$\pm$1.35} & 76.12{\tiny$\pm$1.53} & 70.15{\tiny$\pm$1.34} & 68.74{\tiny$\pm$1.34} & 93.74{\tiny$\pm$1.25} & 75.65{\tiny$\pm$1.35} & 75.49{\tiny$\pm$1.64} \\
ICC(\textsc{gpt-4o}) & 67.58{\tiny$\pm$1.30} & 66.36{\tiny$\pm$1.22} & 65.61{\tiny$\pm$1.12} & 70.15{\tiny$\pm$1.72} & 73.54{\tiny$\pm$1.54} & 80.59{\tiny$\pm$1.28} & 68.61{\tiny$\pm$1.61} & 67.75{\tiny$\pm$1.33} \\
\rowcolor{Blue!10}
ICC (Small)    & 98.25{\tiny$\pm$0.71} & 76.31{\tiny$\pm$1.38} & 85.50{\tiny$\pm$0.78} & 81.75{\tiny$\pm$1.69} & 82.82{\tiny$\pm$1.62} & 89.60{\tiny$\pm$1.52} & 67.89{\tiny$\pm$1.27} & \underline{83.84}{\tiny$\pm$1.53} \\
\rowcolor{Blue!10}
ICC (Medium)   & 98.64{\tiny$\pm$0.58} & \underline{81.02}{\tiny$\pm$1.31} & \underline{93.28}{\tiny$\pm$0.56} & \underline{83.02}{\tiny$\pm$1.69} & \underline{86.04}{\tiny$\pm$1.52} & \underline{95.98}{\tiny$\pm$1.04} & \underline{76.77}{\tiny$\pm$1.39} & 77.18{\tiny$\pm$1.67} \\
\rowcolor{Blue!10}
ICC (Medium:\ref{sec:finetune_image})    & \underline{98.88}{\tiny$\pm$0.55} & 71.39{\tiny$\pm$1.31} & 65.04{\tiny$\pm$1.01} & 72.72{\tiny$\pm$1.37} & 83.04{\tiny$\pm$1.55} & \textbf{96.47}{\tiny$\pm$0.95} & \textbf{78.46}{\tiny$\pm$1.46} & \textbf{86.19}{\tiny$\pm$1.53} \\
\rowcolor{Blue!10}
ICC (Large)    & \textbf{99.52}{\tiny$\pm$0.22} & \textbf{84.29}{\tiny$\pm$1.26} & \textbf{94.43}{\tiny$\pm$0.40} & \textbf{83.72}{\tiny$\pm$1.71} & \textbf{87.27}{\tiny$\pm$1.44} & 94.14{\tiny$\pm$1.26} & 73.42{\tiny$\pm$1.47} & 81.72{\tiny$\pm$1.62} \\
\bottomrule
\end{tabular}
}
\end{sc}
\end{small}
\end{center}
\end{table}

\paragraph{Results.} The quantitative evaluation of different models is summarized in~\Cref{tbl:img_cond} and qualitative examples are shown in~\Cref{sec:appendix_cond}. Similar to results in~\Cref{sec:finetune_image}, zero-shot performance of \textsc{gpt-4o} is promising but ultimately falls short of our finetuned approach. Our finetuned models outperform all baselines on ImageNet and Stanford 40 Action. In general, our method with higher per-image token lengths performs better in this conditional clustering task. Unlike experiments in~\Cref{sec:finetune_image} where the difference between different granularity is small, this task requires more fine-grained information and thus using more tokens to represent images is preferred. K-means and caption-based \ictc{} often fail to capture such details, particularly for attributes like \textsc{color}, \textsc{shape}, and \textsc{pattern}, where our method is more than 10\% higher than all baselines.

Our method generalizes to unseen data and conditions from the Stanford 40 Action dataset. Surprisingly, our model trained solely on clustering objects in ImageNet, achieves the highest accuracy. This suggests that the inductive bias from image-based clustering and the visual-language pretraining enables the model to infer clustering objectives implicitly. We notice that the finetuned models are less competitive on \textsc{mood} and \textsc{location}. We attribute this to the training data (ImageNet-with-Attributes), which emphasizes prominent foreground objects (typically non-human), causing the model to overlook cues from human facial expressions or the background. Scaling our approach to more diverse datasets and clustering conditions could mitigate this bias and further strengthen the model's generalization capabilities.

\section{Conclusion}

In-Context Clustering (ICC) generalizes in-context learning to the unsupervised setting. ICC does not make restrictive similarity assumptions on the input data and enables flexible, text-conditioned clustering objectives through prompting. We find that large LLMs provide strong zero-shot performance on text-encoded numeric data, and further show that this capability can be significantly strengthened for smaller and multimodal models through simple fine-tuning using the NTP loss. Multimodal LLMs enhanced by our proposed finetuning achieve impressive performance on image clustering and text-conditioned image clustering. These findings highlight that LLMs can be effectively used to solve clustering tasks that involve complex semantics and contextual understanding.

While we demonstrate ICC's effectiveness and flexibility, ICC is complementary to classical clustering methods, and has certain limitations that would be exciting to address in future work. For application to larger datasets, it would be particularly promising to scale ICC to longer contexts, which can be computationally expensive for LLMs~\citep{li2024longcontextllmsstrugglelong, liu2023lostmiddlelanguagemodels}. Our experiments with average pooling for image features show promise in reducing token usage, and recent advances such as dynamic context selection~\citep{hao2025omnikv_dynamic_context_selection} and token pruning~\citep{chen2024image_fastv, cao2024madtp} can further address the long-context challenge in future work. Moreover, while visualizing attention provides some insights into the way ICC performs clustering, a theoretical understanding of ICC would be particularly valuable. Emergence of clusters in self-attention have been theoretically studied by \citet{geshkovski2023emergence_clusters}, but under a simplified setting (without multi-head attention, feed-forward layers, and layer normalization). Developing theoretical frameworks to explain and exploit these attention structures remains an important open direction.

\section*{Acknowledgments}
We thank Shikai Qiu, Nate Gruver, Zhe Zeng, Lily Li, and Bayan Bruss for helpful discussions. We are grateful for support from the Institute of Information \& Communications Technology Planning \& Evaluation (IITP) with a grant funded by the Ministry of Science and ICT (MSIT) of the Republic of Korea in connection with the Global AI Frontier Lab International Collaborative Research. (No. RS-2024-00469482 \& RS-2024-00509279), 
NSF CAREER IIS-2145492, NSF CDS\&E-MSS 2134216,
NSF HDR-2118310, BigHat Biosciences, and Capital One.
We are also thankful for NYU IT High Performance Computing resources, services, and staff expertise.

\bibliographystyle{plainnat}
\bibliography{main}

\begin{thebibliography}{52}
\providecommand{\natexlab}[1]{#1}
\providecommand{\url}[1]{\texttt{#1}}
\expandafter\ifx\csname urlstyle\endcsname\relax
  \providecommand{\doi}[1]{doi: #1}\else
  \providecommand{\doi}{doi: \begingroup \urlstyle{rm}\Url}\fi

\bibitem[Achiam et~al.(2023)Achiam, Adler, Agarwal, Ahmad, Akkaya, Aleman, Almeida, Altenschmidt, Altman, Anadkat, Avila, et~al.]{gpt4}
OpenAI~Josh Achiam, Steven Adler, Sandhini Agarwal, Lama Ahmad, Ilge Akkaya, Florencia~Leoni Aleman, Diogo Almeida, Janko Altenschmidt, Sam Altman, Shyamal Anadkat, Red Avila, et~al.
\newblock {GPT-4 Technical Report}.
\newblock \emph{arXiv preprint arXiv:2303.08774}, 2023.

\bibitem[AI@Meta(2024)]{llama3}
AI@Meta.
\newblock {Llama 3 Model Card}.
\newblock 2024.
\newblock URL \url{https://github.com/meta-llama/llama3/blob/main/MODEL_CARD.md}.

\bibitem[Alwassel et~al.(2020)Alwassel, Mahajan, Korbar, Torresani, Ghanem, and Tran]{NEURIPS2020_audio_video_clustering}
Humam Alwassel, Dhruv Mahajan, Bruno Korbar, Lorenzo Torresani, Bernard Ghanem, and Du~Tran.
\newblock {Self-Supervised Learning by Cross-Modal Audio-Video Clustering}.
\newblock \emph{Advances in Neural Information Processing Systems (NeurIPS)}, 2020.

\bibitem[Bai et~al.(2023)Bai, Bai, Chu, Cui, Dang, Deng, Fan, Ge, Han, Huang, Hui, Ji, Li, Lin, Lin, Liu, Liu, Lu, Lu, Ma, Men, Ren, Ren, Tan, Tan, Tu, Wang, Wang, Wang, Wu, Xu, Xu, Yang, Yang, Yang, Yang, Yao, Yu, Yuan, Yuan, Zhang, Zhang, Zhang, Zhang, Zhou, Zhou, Zhou, and Zhu]{qwen}
Jinze Bai, Shuai Bai, Yunfei Chu, Zeyu Cui, Kai Dang, Xiaodong Deng, Yang Fan, Wenbin Ge, Yu~Han, Fei Huang, Binyuan Hui, Luo Ji, Mei Li, Junyang Lin, Runji Lin, Dayiheng Liu, Gao Liu, Chengqiang Lu, Keming Lu, Jianxin Ma, Rui Men, Xingzhang Ren, Xuancheng Ren, Chuanqi Tan, Sinan Tan, Jianhong Tu, Peng Wang, Shijie Wang, Wei Wang, Shengguang Wu, Benfeng Xu, Jin Xu, An~Yang, Hao Yang, Jian Yang, Shusheng Yang, Yang Yao, Bowen Yu, Hongyi Yuan, Zheng Yuan, Jianwei Zhang, Xingxuan Zhang, Yichang Zhang, Zhenru Zhang, Chang Zhou, Jingren Zhou, Xiaohuan Zhou, and Tianhang Zhu.
\newblock Qwen technical report.
\newblock \emph{arXiv preprint arXiv:2309.16609}, 2023.

\bibitem[Brown et~al.(2020)Brown, Mann, Ryder, Subbiah, Kaplan, Dhariwal, Neelakantan, Shyam, Sastry, Askell, Agarwal, Herbert-Voss, Krueger, Henighan, Child, Ramesh, Ziegler, Wu, Winter, Hesse, Chen, Sigler, Litwin, Gray, Chess, Clark, Berner, McCandlish, Radford, Sutskever, and Amodei]{NEURIPS2020_fewshotlearner}
Tom Brown, Benjamin Mann, Nick Ryder, Melanie Subbiah, Jared~D Kaplan, Prafulla Dhariwal, Arvind Neelakantan, Pranav Shyam, Girish Sastry, Amanda Askell, Sandhini Agarwal, Ariel Herbert-Voss, Gretchen Krueger, Tom Henighan, Rewon Child, Aditya Ramesh, Daniel Ziegler, Jeffrey Wu, Clemens Winter, Chris Hesse, Mark Chen, Eric Sigler, Mateusz Litwin, Scott Gray, Benjamin Chess, Jack Clark, Christopher Berner, Sam McCandlish, Alec Radford, Ilya Sutskever, and Dario Amodei.
\newblock {Language Models are Few-Shot Learners}.
\newblock \emph{Advances in Neural Information Processing Systems (NeurIPS)}, 2020.

\bibitem[Bubeck et~al.(2023)Bubeck, Chandrasekaran, Eldan, Gehrke, Horvitz, Kamar, Lee, Lee, Li, Lundberg, et~al.]{sparks}
S{\'e}bastien Bubeck, Varun Chandrasekaran, Ronen Eldan, Johannes Gehrke, Eric Horvitz, Ece Kamar, Peter Lee, Yin~Tat Lee, Yuanzhi Li, Scott Lundberg, et~al.
\newblock {Sparks of artificial general intelligence: Early experiments with gpt-4}.
\newblock \emph{arXiv preprint arXiv:2303.12712}, 2023.

\bibitem[Chang et~al.(2017)Chang, Wang, Meng, Xiang, and Pan]{adaptive_image_clustering_Chang_2017_ICCV}
Jianlong Chang, Lingfeng Wang, Gaofeng Meng, Shiming Xiang, and Chunhong Pan.
\newblock {Deep Adaptive Image Clustering}.
\newblock \emph{International Conference on Computer Vision (ICCV)}, 2017.

\bibitem[Chen et~al.(2024)Chen, Zhao, Liu, Bai, Lin, Zhou, and Chang]{chen2024image_fastv}
Liang Chen, Haozhe Zhao, Tianyu Liu, Shuai Bai, Junyang Lin, Chang Zhou, and Baobao Chang.
\newblock {An Image is Worth 1/2 Tokens After Layer 2: Plug-and-Play Inference Acceleration for Large Vision-Language Models}.
\newblock \emph{European Conference on Computer Vision (ECCV)}, 2024.

\bibitem[Ester et~al.(1996)Ester, Kriegel, Sander, and Xu]{dbscan}
Martin Ester, Hans-Peter Kriegel, J\"{o}rg Sander, and Xiaowei Xu.
\newblock A density-based algorithm for discovering clusters in large spatial databases with noise.
\newblock \emph{International Conference on Knowledge Discovery and Data Mining}, 1996.

\bibitem[Garg et~al.(2022)Garg, Tsipras, Liang, and Valiant]{garg2022llm_simple_function}
Shivam Garg, Dimitris Tsipras, Percy Liang, and Gregory Valiant.
\newblock {What Can Transformers Learn In-Context? A Case Study of Simple Function Classes}.
\newblock \emph{Advances in Neural Information Processing Systems (NeurIPS)}, 2022.

\bibitem[Geshkovski et~al.(2023)Geshkovski, Letrouit, Polyanskiy, and Rigollet]{geshkovski2023emergence_clusters}
Borjan Geshkovski, Cyril Letrouit, Yury Polyanskiy, and Philippe Rigollet.
\newblock The emergence of clusters in self-attention dynamics.
\newblock \emph{Advances in Neural Information Processing Systems (NeurIPS)}, 2023.

\bibitem[Gruver et~al.(2023)Gruver, Finzi, Qiu, and Wilson]{gruver2023llmtime}
Nate Gruver, Marc Finzi, Shikai Qiu, and Andrew~Gordon Wilson.
\newblock {Large Language Models Are Zero Shot Time Series Forecasters}.
\newblock \emph{Advances in Neural Information Processing Systems (NeurIPS)}, 2023.

\bibitem[Gu{\'e}rin and Boots(2018)]{guerin2018improving_image_cnn}
Joris Gu{\'e}rin and Byron Boots.
\newblock {Improving image clustering with multiple pretrained cnn feature extractors}.
\newblock \emph{arXiv preprint arXiv:1807.07760}, 2018.

\bibitem[Guo et~al.(2020)Guo, Codella, Karlinsky, Codella, Smith, Saenko, Rosing, and Feris]{guo2020cdfsl}
Yunhui Guo, Noel~C Codella, Leonid Karlinsky, James~V Codella, John~R Smith, Kate Saenko, Tajana Rosing, and Rogerio Feris.
\newblock A broader study of cross-domain few-shot learning.
\newblock \emph{European Conference on Computer Vision (ECCV)}, 2020.

\bibitem[Hao et~al.(2025)Hao, Zhu, Wang, Yu, Xin, Zheng, Ren, and Guo]{hao2025omnikv_dynamic_context_selection}
Jitai Hao, Yuke Zhu, Tian Wang, Jun Yu, Xin Xin, Bo~Zheng, Zhaochun Ren, and Sheng Guo.
\newblock {Omni{KV}: Dynamic Context Selection for Efficient Long-Context {LLM}s}.
\newblock \emph{International Conference on Learning Representations (ICLR)}, 2025.

\bibitem[Helber et~al.(2019)Helber, Bischke, Dengel, and Borth]{eurosat}
Patrick Helber, Benjamin Bischke, Andreas Dengel, and Damian Borth.
\newblock {EuroSAT: A Novel Dataset and Deep Learning Benchmark for Land Use and Land Cover Classification}.
\newblock \emph{IEEE Journal of Selected Topics in Applied Earth Observations and Remote Sensing}, 2019.

\bibitem[Hu et~al.(2021)Hu, Shen, Wallis, Allen-Zhu, Li, Wang, Wang, and Chen]{hu2021lora}
Edward~J Hu, Yelong Shen, Phillip Wallis, Zeyuan Allen-Zhu, Yuanzhi Li, Shean Wang, Lu~Wang, and Weizhu Chen.
\newblock Lora: Low-rank adaptation of large language models.
\newblock \emph{arXiv preprint arXiv:2106.09685}, 2021.

\bibitem[Huang and Chang(2023)]{huang-chang-2023-llm-survey}
Jie Huang and Kevin Chen-Chuan Chang.
\newblock Towards reasoning in large language models: A survey.
\newblock \emph{Findings of the Association for Computational Linguistics}, 2023.

\bibitem[Ikotun et~al.(2023)Ikotun, Ezugwu, Abualigah, Abuhaija, and Heming]{kmeans-review}
Abiodun~M. Ikotun, Absalom~E. Ezugwu, Laith Abualigah, Belal Abuhaija, and Jia Heming.
\newblock K-means clustering algorithms: A comprehensive review, variants analysis, and advances in the era of big data.
\newblock \emph{Information Sciences}, 2023.

\bibitem[Jain et~al.(1999)Jain, Murty, and Flynn]{traditional_clustering_survey_1999}
A.~K. Jain, M.~N. Murty, and P.~J. Flynn.
\newblock {Data clustering: a review}.
\newblock \emph{ACM Comput. Surv.}, 1999.

\bibitem[Jiang et~al.(2024)Jiang, LI, Zhang, Wu, Luo, Ahn, Han, Abdi, Li, Lin, Yang, and Qiu]{jiang2024minference}
Huiqiang Jiang, YUCHENG LI, Chengruidong Zhang, Qianhui Wu, Xufang Luo, Surin Ahn, Zhenhua Han, Amir~H. Abdi, Dongsheng Li, Chin-Yew Lin, Yuqing Yang, and Lili Qiu.
\newblock {MI}nference 1.0: Accelerating pre-filling for long-context {LLM}s via dynamic sparse attention.
\newblock \emph{Advances in Neural Information Processing Systems (NeurIPS)}, 2024.

\bibitem[Jianjian et~al.(2024)Jianjian, Peng, Shengze, Chong, Yansong, Jiwen, and Tao]{cao2024madtp}
Cao Jianjian, Ye~Peng, Li~Shengze, Yu~Chong, Tang Yansong, Lu~Jiwen, and Chen Tao.
\newblock {MADTP: Multimodal Alignment-Guided Dynamic Token Pruning for Accelerating Vision-Language Transformer}.
\newblock \emph{Conference on Computer Vision and Pattern Recognition (CVPR)}, 2024.

\bibitem[Kwon et~al.(2024)Kwon, Park, Kim, Cho, Ryu, and Lee]{kwon2024imageclusteringtext}
Sehyun Kwon, Jaeseung Park, Minkyu Kim, Jaewoong Cho, Ernest~K. Ryu, and Kangwook Lee.
\newblock Image clustering conditioned on text criteria.
\newblock \emph{International Conference on Learning Representations (ICLR)}, 2024.

\bibitem[Li et~al.(2024{\natexlab{a}})Li, Zhang, Zhang, Zhang, Li, Li, Ma, and Li]{li2024llavanextinterleavetacklingmultiimagevideo}
Feng Li, Renrui Zhang, Hao Zhang, Yuanhan Zhang, Bo~Li, Wei Li, Zejun Ma, and Chunyuan Li.
\newblock Llava-next-interleave: Tackling multi-image, video, and 3d in large multimodal models.
\newblock \emph{arXiv preprint arXiv:2407.07895}, 2024{\natexlab{a}}.

\bibitem[Li et~al.(2024{\natexlab{b}})Li, Zhang, Do, Yue, and Chen]{li2024longcontextllmsstrugglelong}
Tianle Li, Ge~Zhang, Quy~Duc Do, Xiang Yue, and Wenhu Chen.
\newblock Long-context llms struggle with long in-context learning.
\newblock \emph{arXiv preprint arXiv:2404.02060}, 2024{\natexlab{b}}.

\bibitem[Liu et~al.(2024)Liu, Lin, Hewitt, Paranjape, Bevilacqua, Petroni, and Liang]{liu2023lostmiddlelanguagemodels}
Nelson~F. Liu, Kevin Lin, John Hewitt, Ashwin Paranjape, Michele Bevilacqua, Fabio Petroni, and Percy Liang.
\newblock Lost in the middle: How language models use long contexts.
\newblock \emph{Transactions of the Association for Computational Linguistics (ACL)}, 2024.

\bibitem[Liu et~al.(2003)Liu, Liu, Chen, and Ma]{text_clustering}
Tao Liu, Shengping Liu, Zheng Chen, and Wei-Ying Ma.
\newblock An evaluation on feature selection for text clustering.
\newblock \emph{International Conference on Machine Learning (ICML)}, 2003.

\bibitem[Luo et~al.(2025)Luo, An, Zou, Tang, Liu, and Zhang]{luo2025ssdllm}
Yulin Luo, Ruichuan An, Bocheng Zou, Yiming Tang, Jiaming Liu, and Shanghang Zhang.
\newblock Llm as dataset analyst: Subpopulation structure discovery with large language model.
\newblock 2025.

\bibitem[Meinedo and Neto(2003)]{audio_clustering_Meinedo2023}
H.~Meinedo and J.~Neto.
\newblock {Audio segmentation, classification and clustering in a broadcast news task}.
\newblock \emph{International Conference on Acoustics, Speech, and Signal Processing (ICASSP)}, 2003.

\bibitem[Min et~al.(2022)Min, Lewis, Zettlemoyer, and Hajishirzi]{min-etal-2022-metaicl}
Sewon Min, Mike Lewis, Luke Zettlemoyer, and Hannaneh Hajishirzi.
\newblock {M}eta{ICL}: Learning to learn in context.
\newblock 2022.

\bibitem[Mohanty et~al.(2016)Mohanty, Hughes, and Salathé]{plant_disease}
Sharada~P. Mohanty, David~P. Hughes, and Marcel Salathé.
\newblock {Using Deep Learning for Image-Based Plant Disease Detection}.
\newblock \emph{Frontiers in Plant Science}, 2016.

\bibitem[Murtagh and Contreras(2012)]{murtagh2012hierarchical_clustering}
Fionn Murtagh and Pedro Contreras.
\newblock Algorithms for hierarchical clustering: an overview.
\newblock \emph{Wiley Interdisciplinary Reviews: Data Mining and Knowledge Discovery}, 2012.

\bibitem[Najdenkoska et~al.(2023)Najdenkoska, Zhen, and Worring]{najdenkoska2023metavlm}
Ivona Najdenkoska, Xiantong Zhen, and Marcel Worring.
\newblock Meta learning to bridge vision and language models for multimodal few-shot learning.
\newblock 2023.

\bibitem[Nakshatri et~al.(2023)Nakshatri, Liu, Chen, Roth, Goldwasser, and Hopkins]{nakshatri-etal-2023-using}
Nishanth Nakshatri, Siyi Liu, Sihao Chen, Dan Roth, Dan Goldwasser, and Daniel Hopkins.
\newblock {Using LLM for Improving Key Event Discovery: Temporal-Guided News Stream Clustering with Event Summaries}.
\newblock \emph{Findings of the Association for Computational Linguistics: EMNLP}, 2023.

\bibitem[Ng et~al.(2001)Ng, Jordan, and Weiss]{NIPS2001_spectral_clustering}
Andrew Ng, Michael Jordan, and Yair Weiss.
\newblock {On Spectral Clustering: Analysis and an algorithm}.
\newblock \emph{Advances in Neural Information Processing Systems (NeurIPS)}, 2001.

\bibitem[Ravi and Larochelle(2017)]{ravi2017optimizationfewshot}
Sachin Ravi and Hugo Larochelle.
\newblock Optimization as a model for few-shot learning.
\newblock 2017.

\bibitem[Ridnik et~al.(2021)Ridnik, Ben-Baruch, Noy, and Zelnik-Manor]{ridnik2021imagenet21k}
Tal Ridnik, Emanuel Ben-Baruch, Asaf Noy, and Lihi Zelnik-Manor.
\newblock {ImageNet-21K Pretraining for the Masses}.
\newblock \emph{arXiv preprint arXiv:2104.10972}, 2021.

\bibitem[Russakovsky and Fei-Fei(2010)]{RussakovskyECCV10_imagenet_attr}
Olga Russakovsky and Li~Fei-Fei.
\newblock {Attribute Learning in Large-scale Datasets}.
\newblock \emph{ECCV, International Workshop on Parts and Attributes}, 2010.

\bibitem[Shah and Mahajan(2012)]{shah2012document}
Neepa Shah and Sunita Mahajan.
\newblock Document clustering: a detailed review.
\newblock \emph{International Journal of Applied Information Systems}, 2012.

\bibitem[Su et~al.(2024)Su, Zhou, Huang, Li, Wang, Wang, and Xu]{su2024multimodalgeneralizedcategorydiscovery}
Yuchang Su, Renping Zhou, Siyu Huang, Xingjian Li, Tianyang Wang, Ziyue Wang, and Min Xu.
\newblock {Multimodal Generalized Category Discovery}.
\newblock \emph{arXiv preprint arXiv:2409.11624}, 2024.

\bibitem[Tipirneni et~al.(2024)Tipirneni, Adkathimar, Choudhary, Hiranandani, Amjad, Ioannidis, Yuan, and Reddy]{tipirneni2024contextawareclusteringusinglarge}
Sindhu Tipirneni, Ravinarayana Adkathimar, Nurendra Choudhary, Gaurush Hiranandani, Rana~Ali Amjad, Vassilis~N. Ioannidis, Changhe Yuan, and Chandan~K. Reddy.
\newblock {Context-Aware Clustering using Large Language Models}.
\newblock \emph{arXiv preprint arXiv:2405.00988}, 2024.

\bibitem[Tsimpoukelli et~al.(2021)Tsimpoukelli, Menick, Cabi, Eslami, Vinyals, and Hill]{tsimpoukelli2021multimodal_fewshot}
Maria Tsimpoukelli, Jacob Menick, Serkan Cabi, S.~M.~Ali Eslami, Oriol Vinyals, and Felix Hill.
\newblock {Multimodal Few-Shot Learning with Frozen Language Models}.
\newblock \emph{Advances in Neural Information Processing Systems (NeurIPS)}, 2021.

\bibitem[Vacareanu et~al.(2024)Vacareanu, Negru, Suciu, and Surdeanu]{vacareanu2024words}
Robert Vacareanu, Vlad-Andrei Negru, Vasile Suciu, and Mihai Surdeanu.
\newblock {From Words to Numbers: Your Large Language Model Is Secretly A Capable Regressor When Given In-Context Examples}.
\newblock \emph{Conference on Language Modeling (COLM)}, 2024.

\bibitem[Vaswani et~al.(2017)Vaswani, Shazeer, Parmar, Uszkoreit, Jones, Gomez, Kaiser, and Polosukhin]{NIPS2017_attention_all_you_need}
Ashish Vaswani, Noam Shazeer, Niki Parmar, Jakob Uszkoreit, Llion Jones, Aidan~N Gomez, \L~ukasz Kaiser, and Illia Polosukhin.
\newblock Attention is all you need.
\newblock \emph{Advances in Neural Information Processing Systems (NeurIPS)}, 2017.

\bibitem[Viswanathan et~al.(2024)Viswanathan, Gashteovski, Gashteovski, Lawrence, Wu, and Neubig]{viswanathan-etal-2024-fewshot-text-clustering}
Vijay Viswanathan, Kiril Gashteovski, Kiril Gashteovski, Carolin Lawrence, Tongshuang Wu, and Graham Neubig.
\newblock {Large Language Models Enable Few-Shot Clustering}.
\newblock \emph{Transactions of the Association for Computational Linguistics (ACL)}, 2024.

\bibitem[von Luxburg(2007)]{vonluxburg2007tutorialspectralclustering}
Ulrike von Luxburg.
\newblock {A Tutorial on Spectral Clustering}.
\newblock \emph{arXiv preprint arXiv:0711.0189}, 2007.

\bibitem[Ward~Jr(1963)]{ward1963hierarchical}
Joe~H Ward~Jr.
\newblock Hierarchical grouping to optimize an objective function.
\newblock \emph{Journal of the American Statistical Association}, 1963.

\bibitem[Wazarkar and Keshavamurthy(2018)]{image_analysis_clustering_survey_WAZARKAR2018596}
Seema Wazarkar and Bettahally~N. Keshavamurthy.
\newblock {A survey on image data analysis through clustering techniques for real world applications}.
\newblock \emph{Journal of Visual Communication and Image Representation}, 2018.

\bibitem[Yao et~al.(2011)Yao, Jiang, Khosla, Lin, Guibas, and Fei-Fei]{stanford_action}
Bangpeng Yao, Xiaoye Jiang, Aditya Khosla, Andy~Lai Lin, Leonidas Guibas, and Li~Fei-Fei.
\newblock Human action recognition by learning bases of action attributes and parts.
\newblock \emph{International Conference on Computer Vision (ICCV)}, 2011.

\bibitem[Zhai et~al.(2023)Zhai, Mustafa, Kolesnikov, and Beyer]{zhai2023siglip}
Xiaohua Zhai, Basil Mustafa, Alexander Kolesnikov, and Lucas Beyer.
\newblock Sigmoid loss for language image pre-training.
\newblock \emph{International Conference on Computer Vision (ICCV)}, 2023.

\bibitem[Zhang et~al.(2024)Zhang, Mao, Ge, Wang, de~Wynter, Xia, Wu, Song, Lan, and Wei]{zhang2024llmmastermindsurveystrategic}
Yadong Zhang, Shaoguang Mao, Tao Ge, Xun Wang, Adrian de~Wynter, Yan Xia, Wenshan Wu, Ting Song, Man Lan, and Furu Wei.
\newblock {LLM as a Mastermind: A Survey of Strategic Reasoning with Large Language Models}.
\newblock \emph{arXiv preprint arXiv:2404.01230}, 2024.

\bibitem[Zhang et~al.(2023)Zhang, Wang, and Shang]{zhang-etal-2023-clusterllm}
Yuwei Zhang, Zihan Wang, and Jingbo Shang.
\newblock {ClusterLLM: Large Language Models as a Guide for Text Clustering}.
\newblock \emph{Conference on Empirical Methods in Natural Language Processing (EMNLP)}, 2023.

\end{thebibliography}

\clearpage

\appendix

\vbox{
\hsize\textwidth
\linewidth\hsize
\vskip 0.1in
\hrule height 4pt
  \vskip 0.25in
  \vskip -\parskip
\centering
{\LARGE\bf
Appendix
\par}
\vskip 0.29in
  \vskip -\parskip
  \hrule height 1pt
  \vskip 0.09in
}

\section{Additional Results of Numeric Data Clustering}
\label{sec:appendix_numeric}

\begin{figure}[h]
\centering
\includegraphics[width=\textwidth]{images/numeric_zeroshot_t_2c.pdf}
\includegraphics[width=\textwidth]{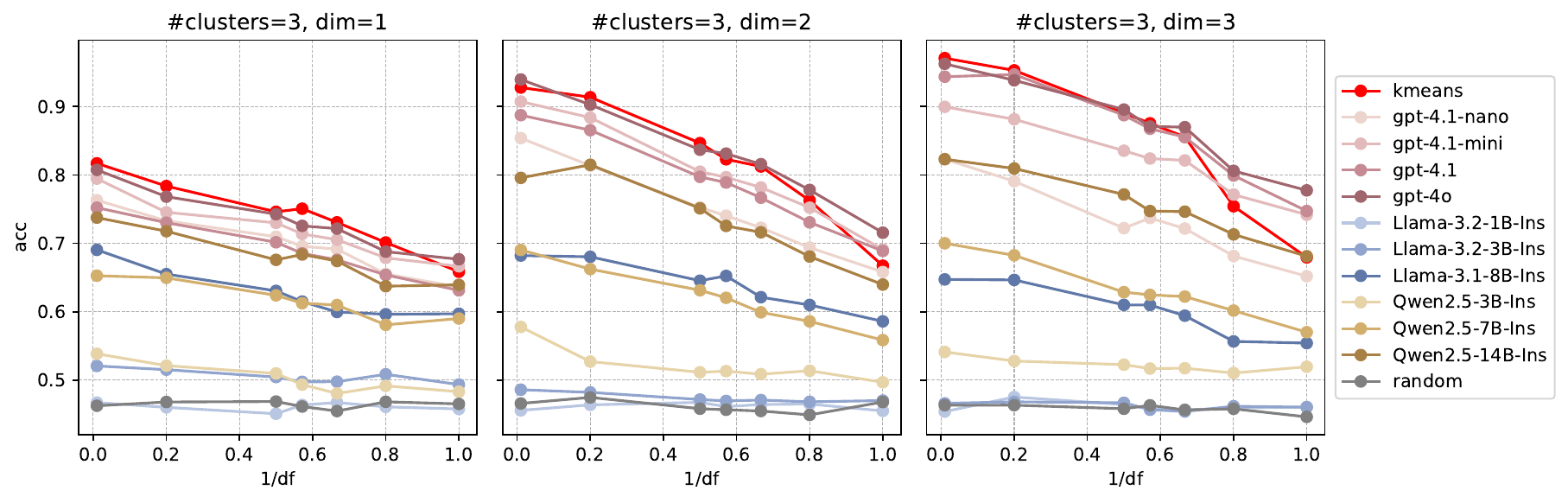}
\includegraphics[width=\textwidth]{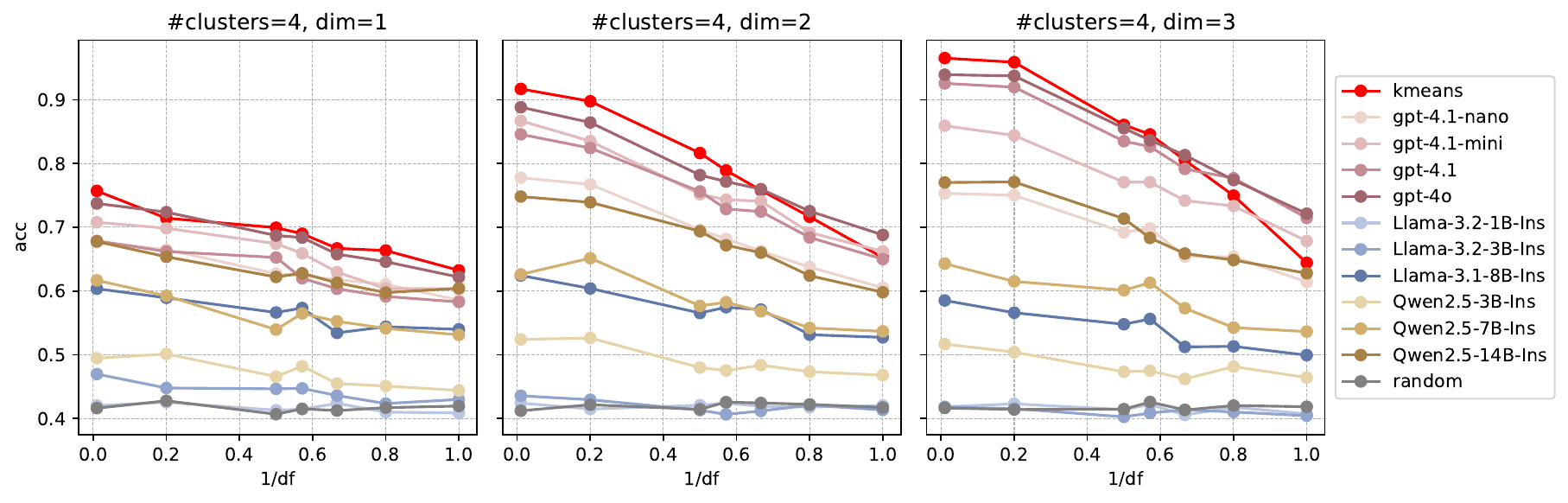}
\caption{Zero-shot Clustering Accuracy. Test data is t-distributed with different degrees of freedom, number of clusters and dimensions. Note that ``Ins'' represents ``Instruct'' in the legend.}
\label{img:zeroshot_full}
\end{figure}

\begin{figure}[t]
\centering
\includegraphics[width=\textwidth]{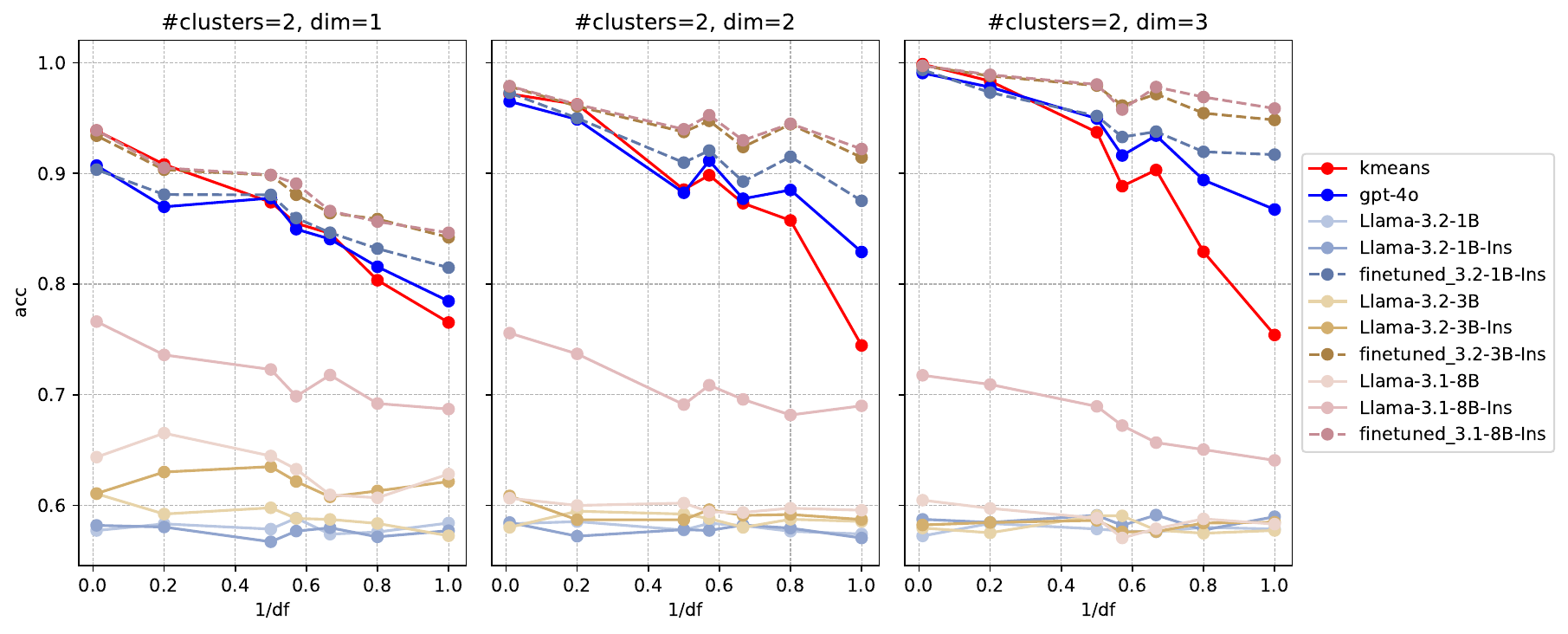}
\includegraphics[width=\textwidth]{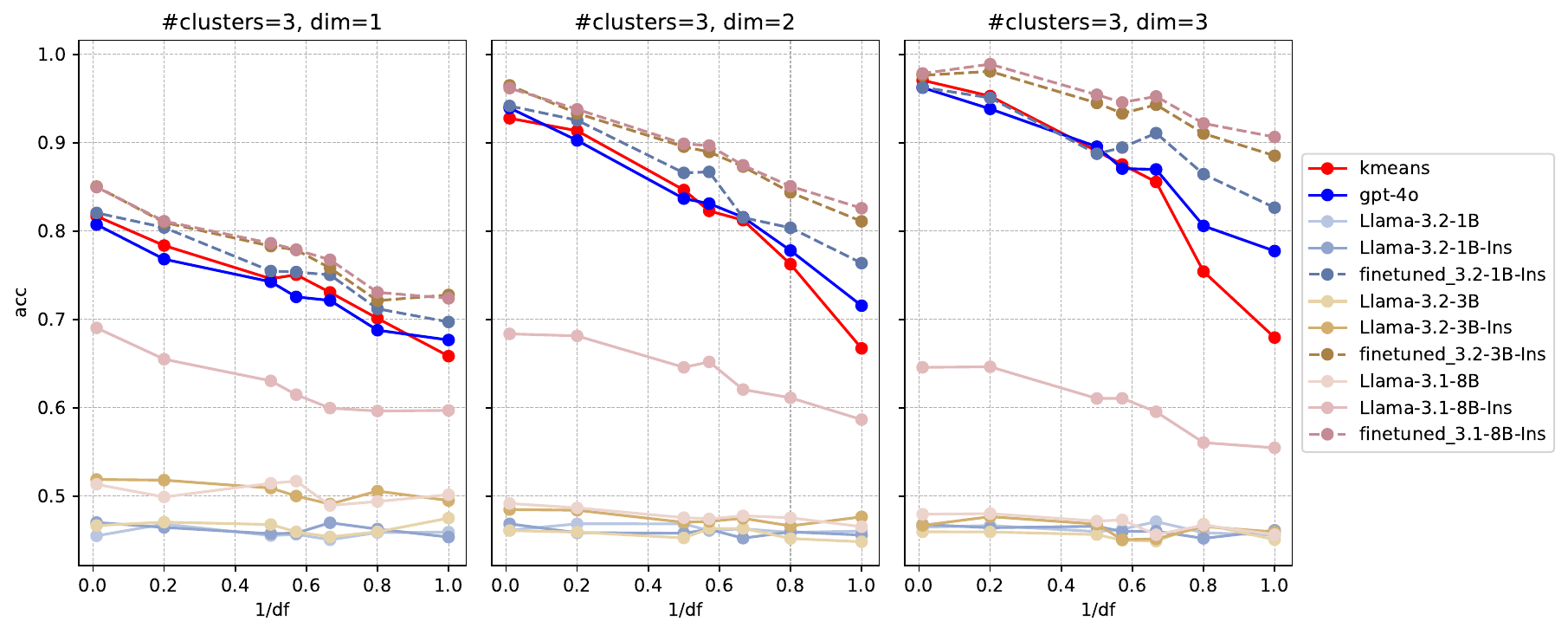}
\includegraphics[width=\textwidth]{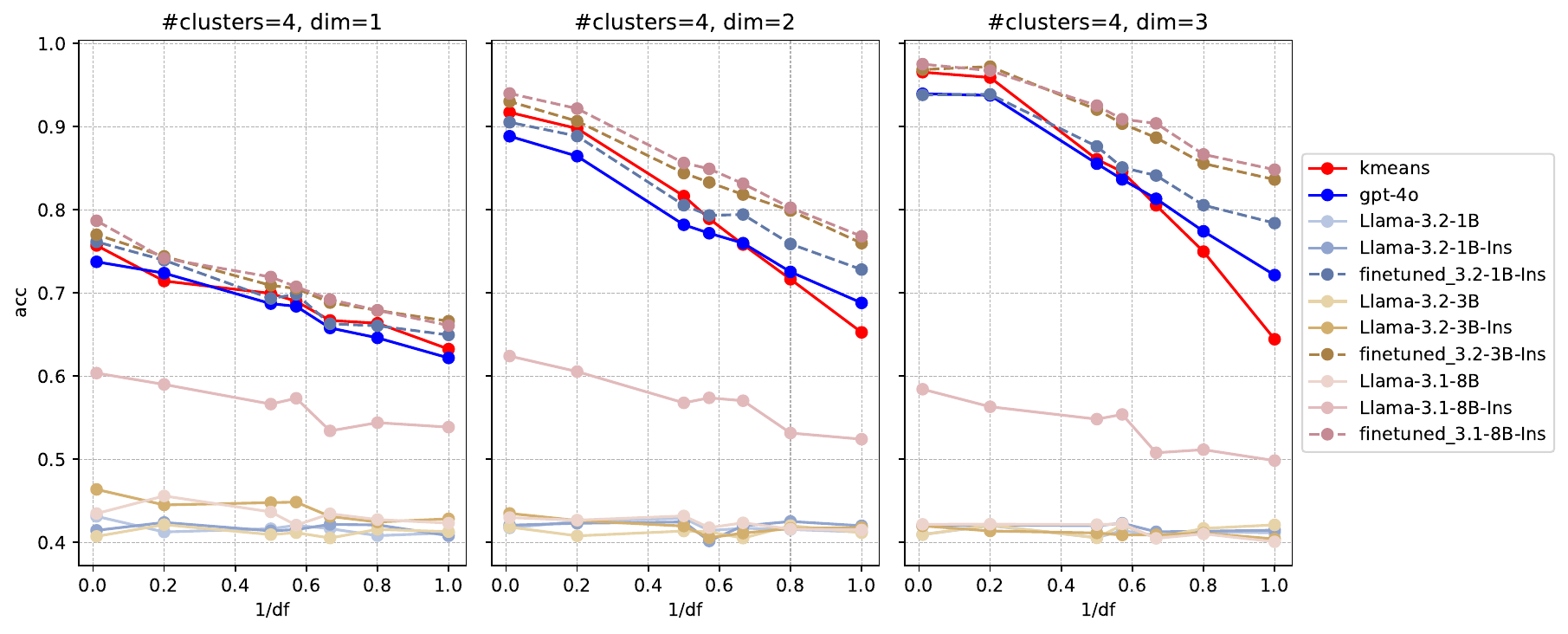}
\caption{Impact of Instruction Tuning and Clustering-Specific Fine-tuning on Clustering Accuracy. Test data is t-distributed with different degrees of freedom, number of clusters and dimensions. Note that ``Ins'' represents ``Instruct'', and ``finetune'' refers to the fine-tuning on t-distributed clustering data with $df \in \{1,2,5,100\}$ as in~\Cref{sec:finetune_num}.}
\label{img:finetune_instruct}
\end{figure}

\clearpage

\begin{table}[h]
\caption{Average Clustering Accuracy with One Standard Error on Lognormal Data. \textsc{finetuned} represents the fine-tuned \textsc{llama-3.1-8b} model on t-distributed clustering data with $df \in \{1,2,5,100\}$ as in~\Cref{sec:finetune_num}. Although the model is not fine-tuned on lognormal data, it still outperforms other models in almost all settings. }
\label{tbl:lognormal}
\begin{center}
\begin{small}
\begin{sc}
\begin{tabular}{llccc}
\toprule
    &                       & $c=2$           & $c=3$           & $c=4$           \\
\midrule
\multirow{3}{*}{$dim=1$}
& kmeans                 & 0.86\tiny{$\pm$0.03} & 0.77\tiny{$\pm$0.02} & 0.74\tiny{$\pm$0.02} \\
& gpt-4o                 & 0.87\tiny{$\pm$0.02} & 0.75\tiny{$\pm$0.02} & 0.73\tiny{$\pm$0.02} \\
& finetuned  & \textbf{0.89}\tiny{$\pm$0.02} & \textbf{0.79}\tiny{$\pm$0.02} & \textbf{0.76}\tiny{$\pm$0.02} \\
\midrule
\multirow{3}{*}{$dim=2$}
& kmeans                 & 0.91\tiny{$\pm$0.03} & 0.87\tiny{$\pm$0.02} & 0.82\tiny{$\pm$0.02} \\
& gpt-4o                 & 0.91\tiny{$\pm$0.02} & 0.84\tiny{$\pm$0.02} & 0.80\tiny{$\pm$0.02} \\
& finetuned  & \textbf{0.94}\tiny{$\pm$0.02} & \textbf{0.91}\tiny{$\pm$0.02} & \textbf{0.86}\tiny{$\pm$0.02} \\
\midrule
\multirow{3}{*}{$dim=3$}
& kmeans                 & \textbf{0.98}\tiny{$\pm$0.01} & 0.92\tiny{$\pm$0.02} & 0.91\tiny{$\pm$0.02} \\
& gpt-4o                 & 0.94\tiny{$\pm$0.01} & 0.86\tiny{$\pm$0.02} & 0.88\tiny{$\pm$0.02} \\
& finetuned  & 0.94\tiny{$\pm$0.02} & \textbf{0.94}\tiny{$\pm$0.02} & \textbf{0.92}\tiny{$\pm$0.02} \\
\bottomrule
\end{tabular}
\end{sc}
\end{small}
\end{center}
\end{table}

\begin{table}[ht]
\caption{Sensitivity to Input Order. The reported values are average accuracy on t-distributed (c=2, dim=3) data, with average standard deviation over five runs of permuted input data in parentheses. We use the standard deviation to reflect the consistency of clustering methods given permutations of input data. \textsc{finetuned} denotes the \textsc{llama-3.1-8b} model finetuned on t-distributed clustering data in~\Cref{sec:finetune_num}, and \textsc{finetuned-aug} denotes finetuning on augmented data with 3 times of permutation. We notice that the model with higher clustering accuracy tends to be more invariant to permutation in input data. Data augmentation is also effective in improving the consistency.}
\label{tbl:permutation_invariance}
\begin{center}
\begin{small}
\begin{sc}
\begin{tabular}{lcccc}
\toprule
& df=1 & df=2 & df=5 & df=100 \\
\midrule
k-means & 0.75(0.04) & 0.95(0.03) & \textbf{0.99(0.00)} & \textbf{0.99(0.00)} \\
gpt-4o & 0.83(0.08) & 0.95(0.03) & 0.97(0.02) & 0.98(0.01) \\
finetuned & 0.92(0.04) & 0.97(0.02) & 0.98(0.01) & 0.99(0.01) \\
finetuned-aug & \textbf{0.93(0.03)} & \textbf{0.98(0.01)} & 0.98(0.01) & \textbf{0.99(0.00)} \\
\bottomrule
\end{tabular}
\end{sc}
\end{small}
\end{center}
\end{table}

\clearpage

\section{Emergence of Clusters in Attention}
\label{sec:appendix_attention_clustering}

\subsection{Attention of Different Layers and Attention Heads}
\label{sec:appendix_attention}

\;
\begin{wrapfigure}{r}{0.2\textwidth}
  \vspace{-4.5cm}
  \begin{center}
  \includegraphics[width=0.2\textwidth]{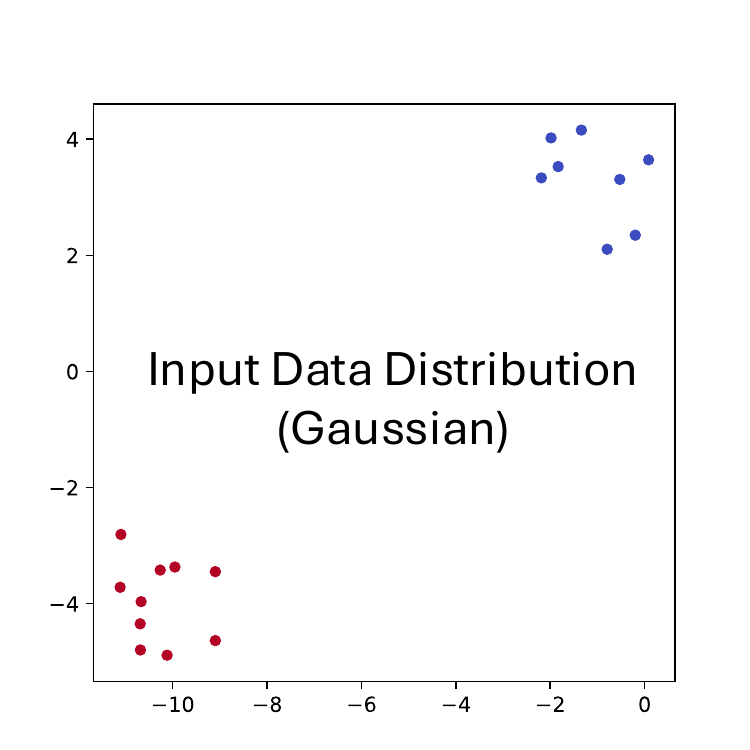}
  \end{center}
\end{wrapfigure}

\begin{figure}[h]
\centering
\vspace{-1cm}
\includegraphics[width=0.8\textwidth]{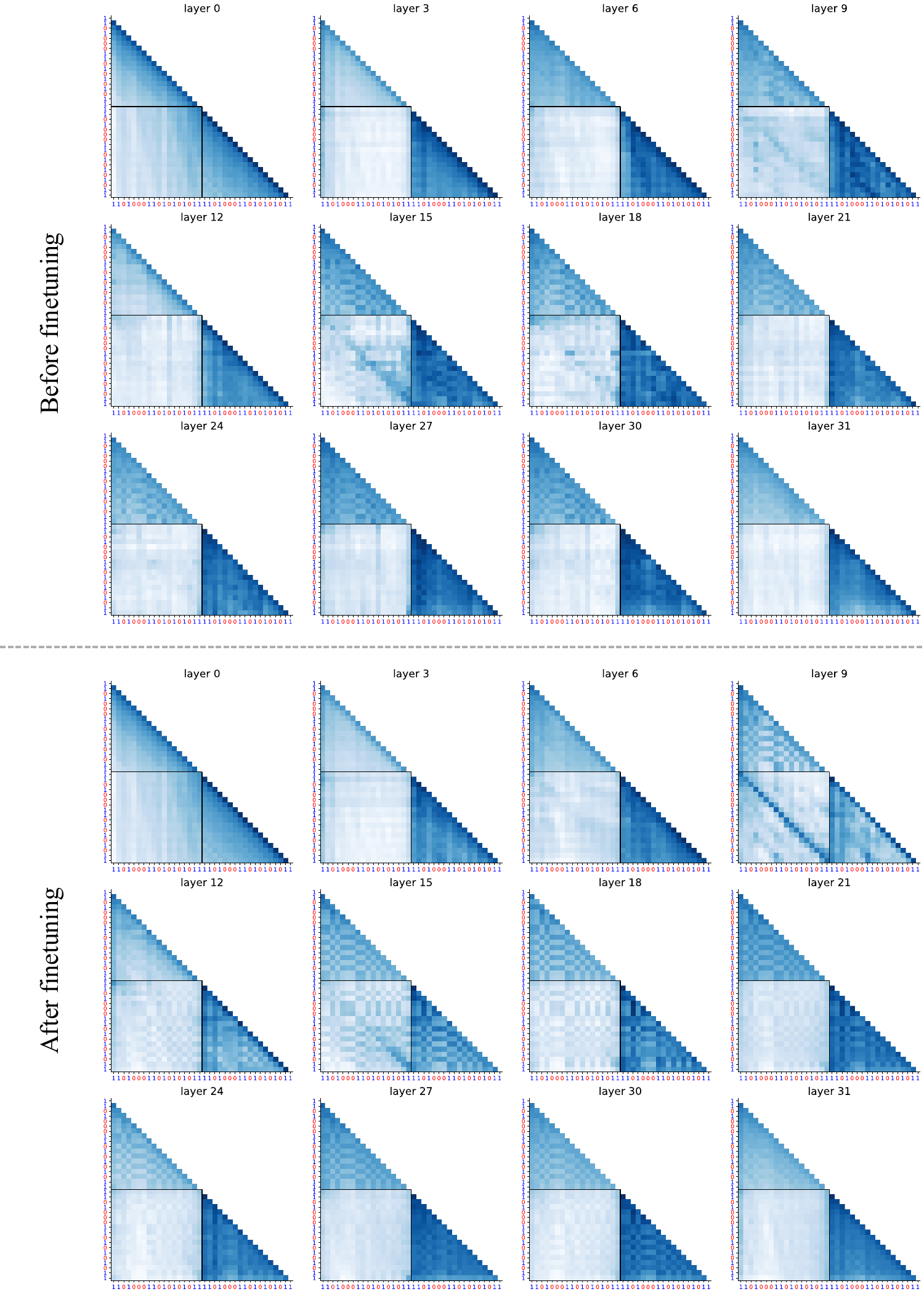}
\caption{Attention Allocation of \textsc{Llama-3.1-8b-Instruct} across Layers. The attention scores are logarithmized for better visualization. Each cluster is generated from a Gaussian distribution, as shown in top right. \Cref{img:attention} is a zoom-in view of layer 15 here.}
\label{img:attention_finetune_morelayers}
\vspace{-2cm}
\end{figure}

\begin{figure}[h]
\centering
\includegraphics[width=0.9\textwidth]{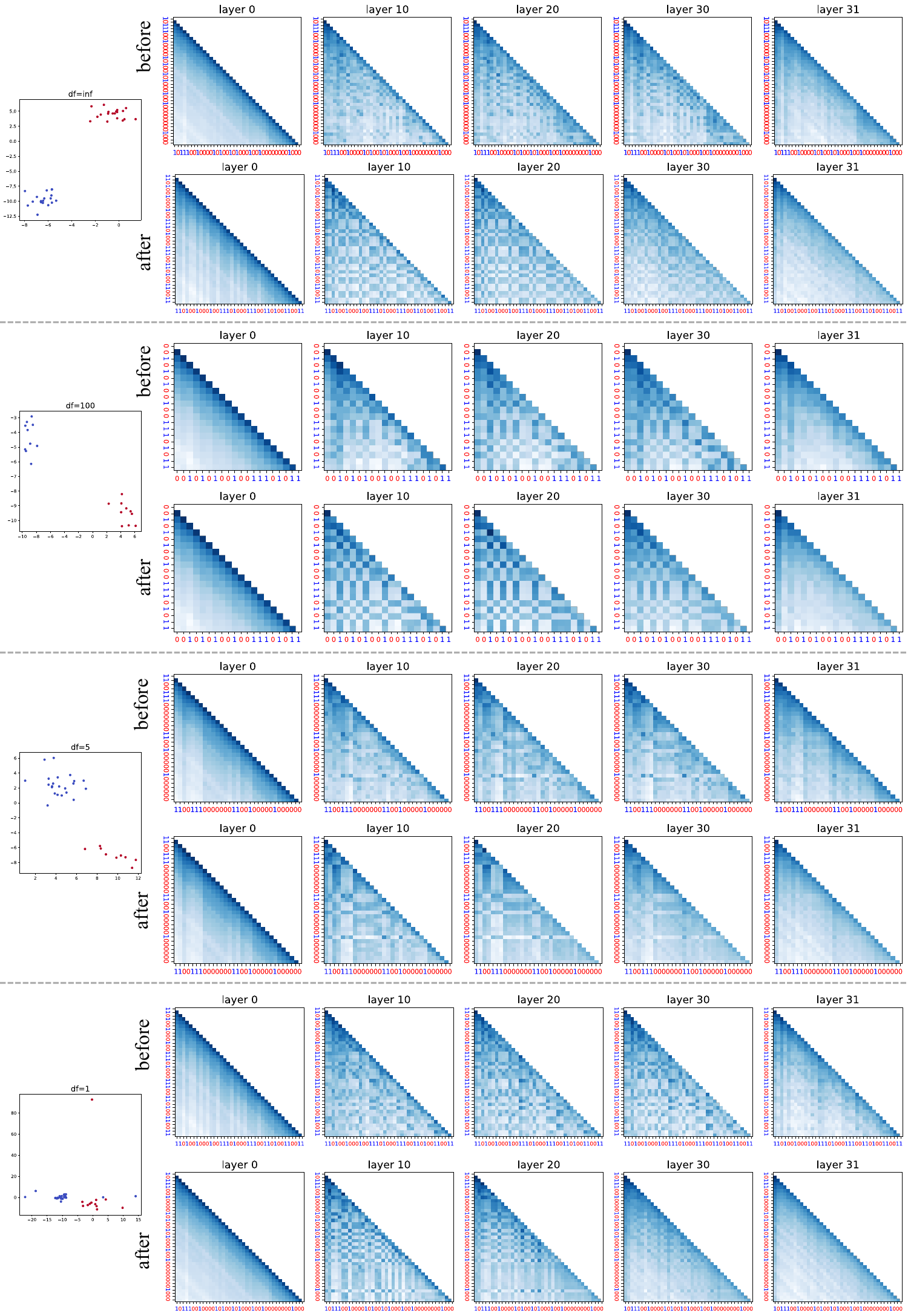}
\caption{Attention Allocation of \textsc{Llama-3.1-8b-Instruct} on $t$-Distributed Data with Different~$df$, before and after Finetuning. Note that $t$-distribution with $df=inf$ is Gaussian. The attention scores are logarithmized for better visualization.}
\label{img:attention_more}
\end{figure}

\begin{figure}[h]
\centering
\includegraphics[width=0.95\textwidth]{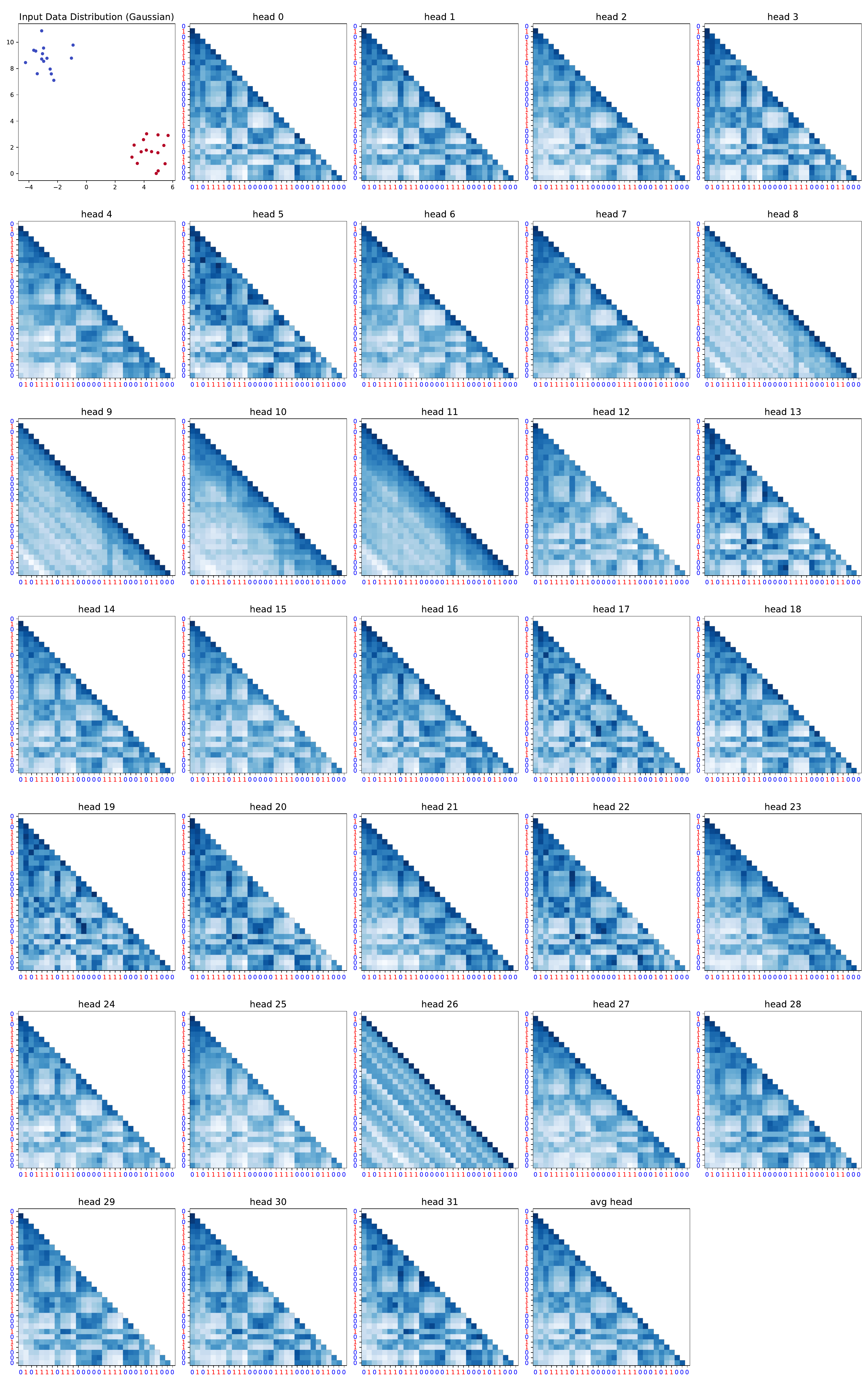}
\caption{Attention Allocation of \textsc{Llama-3.1-8b-Instruct} across attention heads at layer 15. The attention scores are logarithmized for better visualization. Each cluster is generated from a Gaussian distribution, as shown in top left.}
\label{img:attention_heads}
\end{figure}

\clearpage
\subsection{Spectral Clustering}
\label{sec:appendix_spectral}
\begin{figure}[h]
\centering
\includegraphics[width=\textwidth]{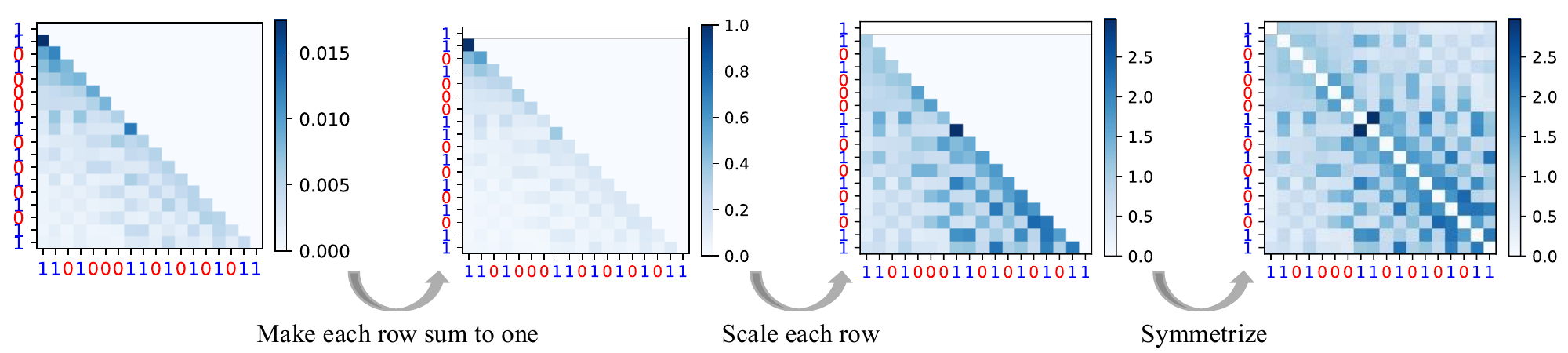}
\caption{Preprocessing Attention Matrix for Spectral Clustering.}
\label{img:attention_spectral}
\end{figure}

\looseness=-100000 As described in~\Cref{sec:attention}, we perform spectral clustering using the input-input attention score matrix $A^{II}$. We first standardize $A^{II}$ so that each row sums to one. Due to causality, early tokens cannot attend to later tokens, making the attention scores scale uneven across rows. For example, the second data point always allocates very high attention to the first one regardless of its semantic similarity. To mitigate this imbalance, we further rescale each row by the number of non-zero entries in the row. Finally, we symmetrize the matrix and the resulting matrix is used as the precomputed affinity matrix for spectral clustering. The complete preprocessing procedure is visualized in~\Cref{img:attention_spectral}. We use the \texttt{ sklearn.cluster.SpectralClustering} implementation. 

\begin{table}[h]
\caption{Spectral Clustering using Attention Scores. Reported values are average accuracy on t-distributed test data as in~\Cref{sec:zeroshot}, with one standard error. Models used here are pretrained \textsc{Llama-3.1-8b-Instruct} and its fine-tuned checkpoint as in~\Cref{sec:finetune_num}. \textsc{SC} represents spectral clustering using attention scores with \textsc{opt} denoting the highest accuracy across all layers and \textsc{l23} denoting the accuracy using a fixed layer 23 (indexing from 0). \textsc{Gen} represents generation using direct LLM prompting. Spectral clustering using attention achieves surprisingly competitive performance that outperforms the raw generation before finetuning.}
\label{tbl:spec_acc}
\begin{center}
\begin{small}
\begin{sc}
\resizebox{0.9\textwidth}{!}{
\begin{tabular}{ccccccccc}
\toprule
  model   & method & df=1       &  df=1.25 &  df=1.5       & df=1.75       & df=2      & df=5       &  df=100       \\
\midrule
\multicolumn{2}{c}{\emph{num of clusters = 2, dim = 1}} \\

\multirow{3}{4em}{pretrained}  & SC(opt) & 0.68\tiny{$\pm$0.01} & 0.70\tiny{$\pm$0.01} & 0.73\tiny{$\pm$0.01} & 0.73\tiny{$\pm$0.02} & 0.71\tiny{$\pm$0.01} & 0.79\tiny{$\pm$0.02} & 0.79\tiny{$\pm$0.02} \\

 & SC(l23) & 0.68\tiny{$\pm$0.01} & 0.68\tiny{$\pm$0.01} & 0.72\tiny{$\pm$0.01} & 0.73\tiny{$\pm$0.02} & 0.71\tiny{$\pm$0.02} & 0.79\tiny{$\pm$0.02} & 0.79\tiny{$\pm$0.02} \\
 & Gen & 0.69\tiny{$\pm$0.01} & 0.69\tiny{$\pm$0.01} & 0.72\tiny{$\pm$0.01} & 0.70\tiny{$\pm$0.01} & 0.72\tiny{$\pm$0.01} & 0.74\tiny{$\pm$0.02} & 0.77\tiny{$\pm$0.01} \\

\arrayrulecolor{black!30}\midrule

\multirow{3}{4em}{finetuned} & SC(opt) & 0.70\tiny{$\pm$0.01} & 0.72\tiny{$\pm$0.01} & 0.73\tiny{$\pm$0.01} & 0.74\tiny{$\pm$0.02} & 0.74\tiny{$\pm$0.02} & 0.79\tiny{$\pm$0.02} & 0.79\tiny{$\pm$0.02} \\

 & SC(l23) & 0.67\tiny{$\pm$0.01} & 0.70\tiny{$\pm$0.02} & 0.72\tiny{$\pm$0.02} & 0.72\tiny{$\pm$0.02} & 0.72\tiny{$\pm$0.02} & 0.76\tiny{$\pm$0.02} & 0.75\tiny{$\pm$0.02} \\
 & Gen & 0.85\tiny{$\pm$0.01} & 0.86\tiny{$\pm$0.01} & 0.87\tiny{$\pm$0.01} & 0.89\tiny{$\pm$0.01} & 0.90\tiny{$\pm$0.01} & 0.91\tiny{$\pm$0.01} & 0.94\tiny{$\pm$0.01} \\

\arrayrulecolor{black}\midrule

\multicolumn{2}{c}{\emph{num of clusters = 2, dim = 2}} \\

\multirow{3}{4em}{pretrained}  & SC(opt) & 0.75\tiny{$\pm$0.01} & 0.76\tiny{$\pm$0.02} & 0.79\tiny{$\pm$0.02} & 0.78\tiny{$\pm$0.02} & 0.81\tiny{$\pm$0.02} & 0.82\tiny{$\pm$0.02} & 0.88\tiny{$\pm$0.02} \\

& SC(l23) & 0.71\tiny{$\pm$0.01} & 0.74\tiny{$\pm$0.02} & 0.73\tiny{$\pm$0.02} & 0.76\tiny{$\pm$0.02} & 0.78\tiny{$\pm$0.02} & 0.80\tiny{$\pm$0.02} & 0.87\tiny{$\pm$0.02} \\

& Gen & 0.69\tiny{$\pm$0.01} & 0.68\tiny{$\pm$0.01} & 0.69\tiny{$\pm$0.01} & 0.71\tiny{$\pm$0.01} & 0.69\tiny{$\pm$0.01} & 0.74\tiny{$\pm$0.02} & 0.75\tiny{$\pm$0.01} \\

\arrayrulecolor{black!30}\midrule

\multirow{3}{4em}{finetuned}  & SC(opt) & 0.84\tiny{$\pm$0.01} & 0.84\tiny{$\pm$0.02} & 0.85\tiny{$\pm$0.02} & 0.87\tiny{$\pm$0.01} & 0.87\tiny{$\pm$0.01} & 0.89\tiny{$\pm$0.02} & 0.96\tiny{$\pm$0.01} \\

& SC(l23) & 0.77\tiny{$\pm$0.02} & 0.81\tiny{$\pm$0.02} & 0.80\tiny{$\pm$0.02} & 0.82\tiny{$\pm$0.02} & 0.83\tiny{$\pm$0.02} & 0.87\tiny{$\pm$0.02} & 0.94\tiny{$\pm$0.01} \\

& Gen & 0.92\tiny{$\pm$0.01} & 0.94\tiny{$\pm$0.01} & 0.93\tiny{$\pm$0.01} & 0.95\tiny{$\pm$0.01} & 0.94\tiny{$\pm$0.01} & 0.96\tiny{$\pm$0.01} & 0.98\tiny{$\pm$0.01}  \\

\arrayrulecolor{black}\midrule

\multicolumn{2}{c}{\emph{num of clusters = 2, dim = 3}} \\

\multirow{3}{4em}{pretrained}  & SC(opt)  & 0.77\tiny{$\pm$0.02} & 0.79\tiny{$\pm$0.02} & 0.78\tiny{$\pm$0.02} & 0.80\tiny{$\pm$0.02} & 0.83\tiny{$\pm$0.02} & 0.85\tiny{$\pm$0.02} & 0.88\tiny{$\pm$0.02} \\

& SC(l23)  & 0.68\tiny{$\pm$0.01} & 0.71\tiny{$\pm$0.02} & 0.73\tiny{$\pm$0.02} & 0.74\tiny{$\pm$0.02} & 0.76\tiny{$\pm$0.02} & 0.81\tiny{$\pm$0.02} & 0.85\tiny{$\pm$0.02} \\

& Gen & 0.64\tiny{$\pm$0.01} & 0.65\tiny{$\pm$0.01} & 0.66\tiny{$\pm$0.01} & 0.67\tiny{$\pm$0.01} & 0.69\tiny{$\pm$0.01} & 0.70\tiny{$\pm$0.02} & 0.71\tiny{$\pm$0.02} \\

\arrayrulecolor{black!30}\midrule

\multirow{3}{4em}{finetuned}  & SC(opt) & 0.90\tiny{$\pm$0.01} & 0.91\tiny{$\pm$0.01} & 0.93\tiny{$\pm$0.01} & 0.91\tiny{$\pm$0.01} & 0.93\tiny{$\pm$0.01} & 0.96\tiny{$\pm$0.01} & 0.99\tiny{$\pm$0.00} \\

& SC(l23) & 0.83\tiny{$\pm$0.02} & 0.86\tiny{$\pm$0.02} & 0.89\tiny{$\pm$0.02} & 0.87\tiny{$\pm$0.02} & 0.91\tiny{$\pm$0.01} & 0.95\tiny{$\pm$0.01} & 0.97\tiny{$\pm$0.01} \\

& Gen & 0.96\tiny{$\pm$0.01} & 0.97\tiny{$\pm$0.01} & 0.98\tiny{$\pm$0.00} & 0.96\tiny{$\pm$0.01} & 0.98\tiny{$\pm$0.00} & 0.99\tiny{$\pm$0.00} & 1.00\tiny{$\pm$0.00}\\

\arrayrulecolor{black}\bottomrule
\end{tabular}
}
\vspace{-0.2in}
\end{sc}
\end{small}
\end{center}
\end{table}

\clearpage
\section{Additional Experiment Details and Results of Image Clustering}
\label{sec:appendix_image}

\subsection{Pooling}
\label{sec:appendix_pooling}

\begin{table}[h]
\caption{Pooling kernel size and corresponding per-image token length. The original pixel size is 384x384 with a patch size of 14, resulting in 27x27(729) image tokens. }
\vspace{-0.2in}
\label{tbl:pool}
\begin{center}
\begin{small}
\begin{sc}
\begin{tabular}{ccccccc}
\toprule
    & pooling kernel & token length \\
\midrule
Default   & 1x1 & 27 x 27 (729) \\
Large     & 2x2 & 13 x 13 (169) \\
Medium    & 3x3 & 9 x 9 (81) \\
Small     & 9x9 & 3 x 3 (9) \\        
\bottomrule
\end{tabular}
\end{sc}
\end{small}
\end{center}
\vspace{-0.2in}
\end{table}

\subsection{Out-of-Domain Image datasets}
\label{sec:cdfsl}

To test the generalization capability of the model, we include two more image datasets from Cross-Domain Few-Shot Learning (CD-FSL) Benchmark~\cite{guo2020cdfsl}.

\begin{itemize}
    \item Plant Disease~\cite{plant_disease}: Leaves of different trees that are healthy or have different crop diseases. We construct 100 clustering samples based on the plant names, where each sample contains 10-30 images from 3 random classes.
    \item EuroSAT~\cite{eurosat}: Satellite images of different land use and land cover classes. We construct 100 clustering samples where each sample contains 10-30 images from 3 random classes.
\end{itemize}

\begin{figure}[h]
\centering
\includegraphics[width=0.8\textwidth]{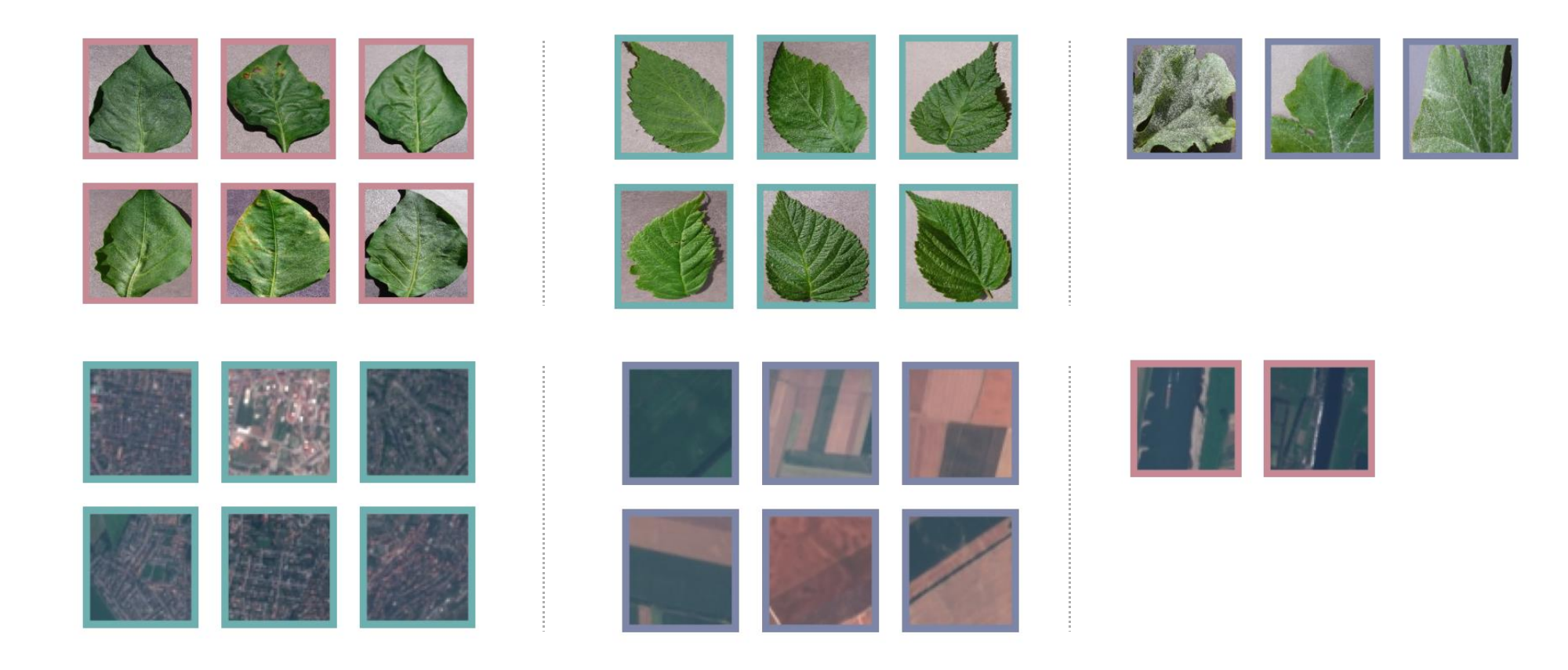}
\caption{Example of Plant Disease and EuroSAT datasets. The color of frame represents different clusters predicted by our model. Our model can generalize to these images that are quite different from ImageNet.}
\label{img:plant}
\end{figure}

\subsection{Attention}

\begin{figure}[h]
\centering
\includegraphics[width=0.9\textwidth]{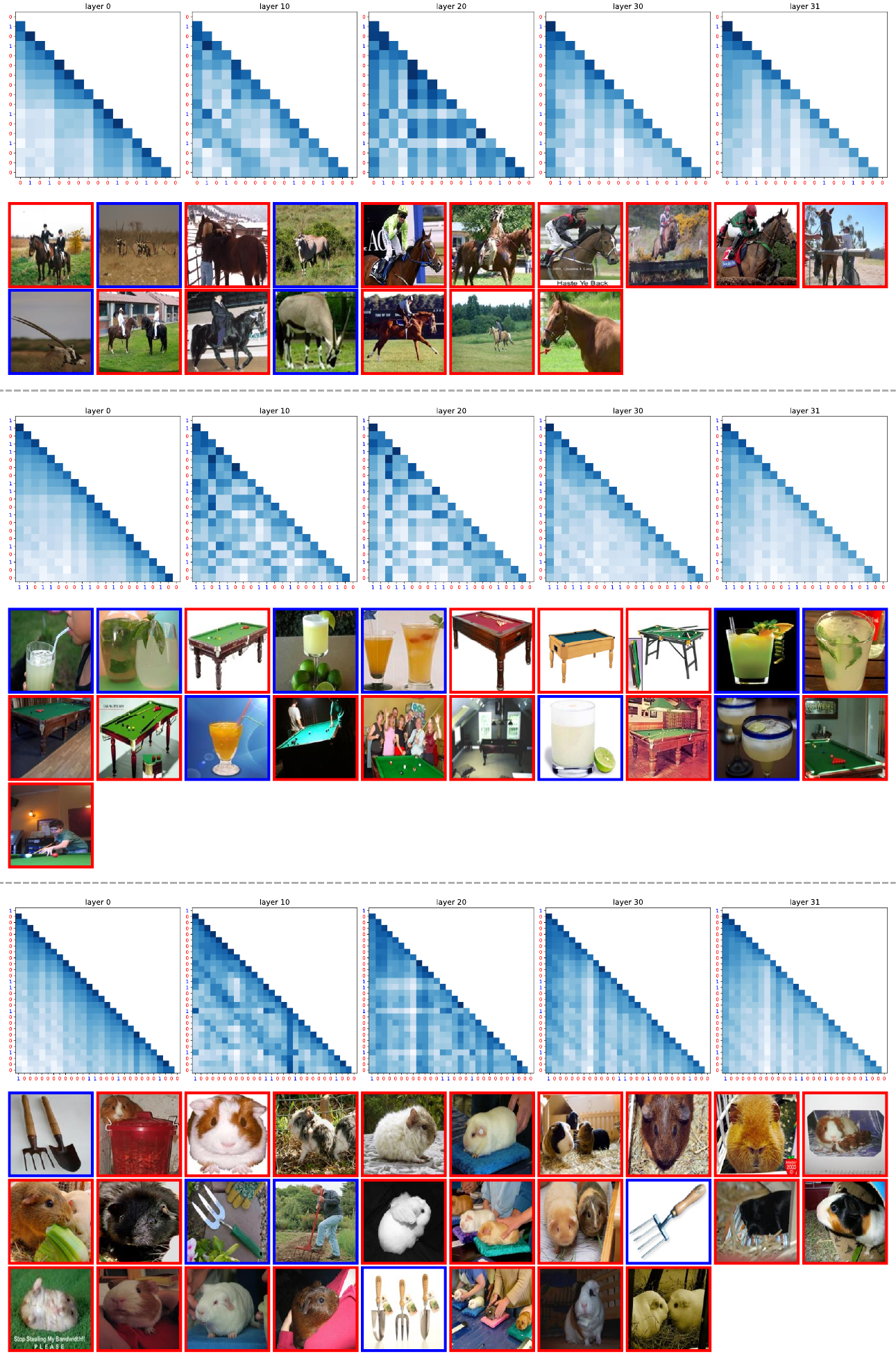}
\caption{Attention Allocation of Image Clustering. Different colors represent different clusters.}
\label{img:image_attention}
\end{figure}

Similar as the numeric experiments in~\Cref{sec:attention}, we visualize the attention allocation for image clustering below (\Cref{img:image_attention}). The model used here is fine-tuned model (medium) as in~\Cref{sec:finetune_image}. The attention scores have block structures that roughly align with the ground-truth identities in intermediate layers. We notice that the allocation of attention weights can be uneven within one cluster, where representative samples are assigned with higher weights. The attention patterns for images are generally more complicated than those for synthetic low-dimensional data due to the semantically rich information in images.

\clearpage
\section{Additional Results for Conditional Image Clustering}
\label{sec:appendix_cond}

\begin{figure}[h]
\centering
\includegraphics[width=0.9\textwidth]{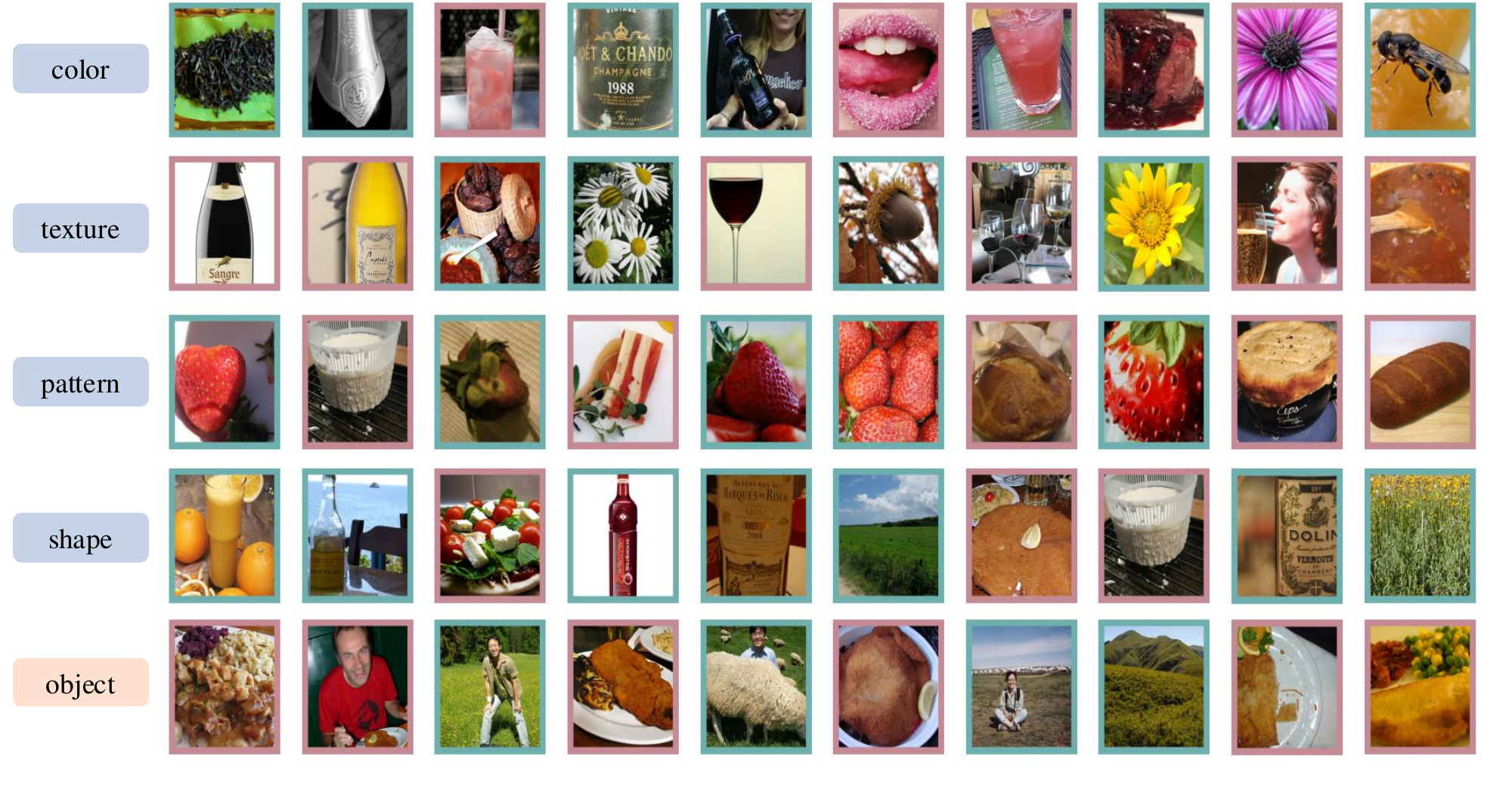}
\includegraphics[width=0.9\textwidth]{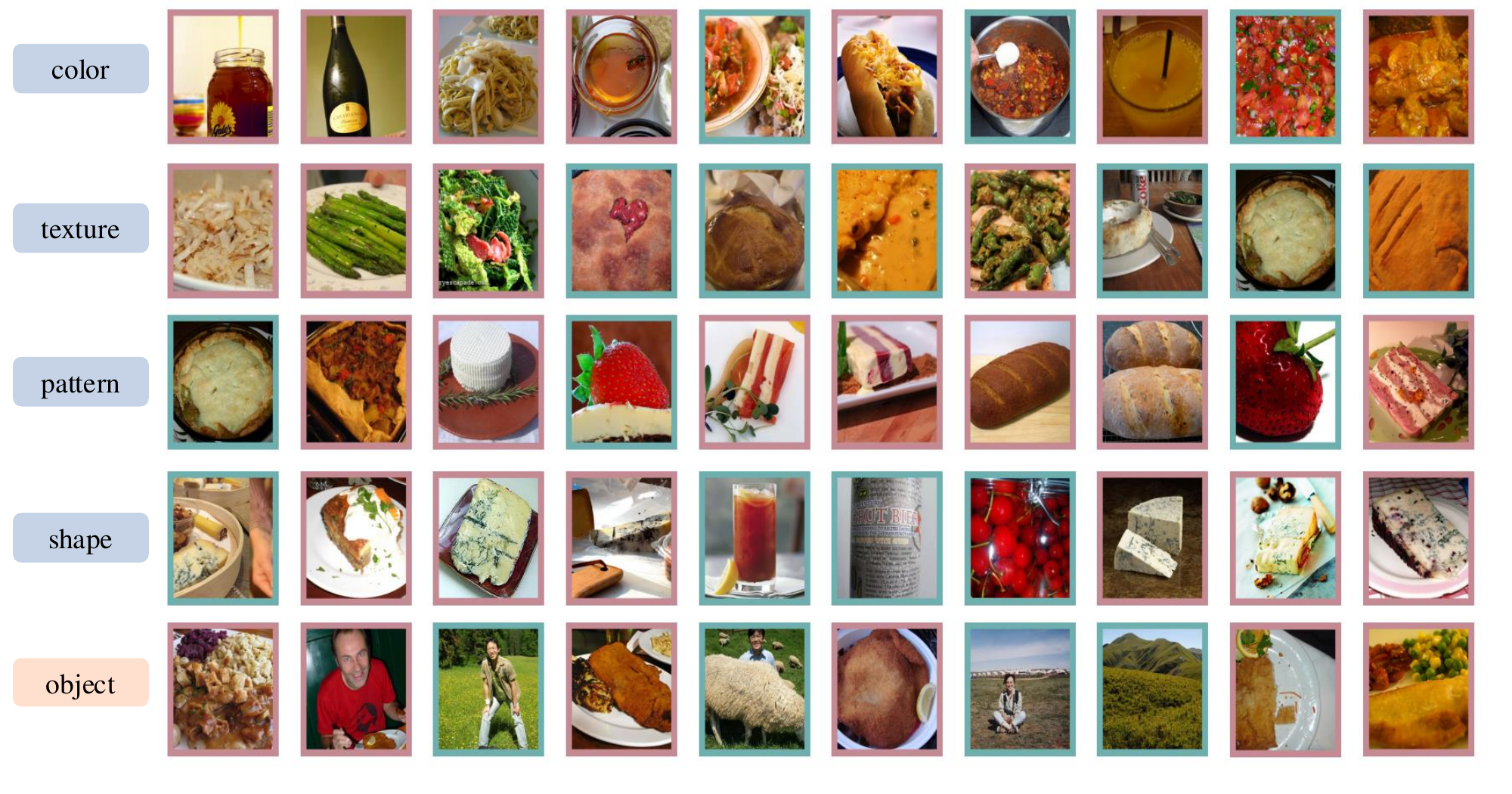}
\caption{Examples of ICC on ImageNet-with-Attributes. The color of the frame indicates different clusters predicted by our model. Most of the images contain multiple objects, making the task more challenging.}
\label{img:img_cond_examples}
\end{figure}

\clearpage

\end{document}